\documentclass[twoside]{article}

\usepackage[utf8]{inputenc} 
\usepackage[T1]{fontenc}    

\usepackage{amsmath}
\usepackage{amssymb}
\usepackage{graphicx}
\usepackage{natbib}
\usepackage{hyperref}       
\usepackage{url}            
\usepackage{booktabs}       
\usepackage{amsfonts}       
\usepackage{nicefrac}       
\usepackage{microtype}      
\usepackage{xcolor}         
\usepackage{algorithm}
\usepackage{algorithmic}
\usepackage{xkcdcolors}
\usepackage{subfigure}
\usepackage{wrapfig}
\usepackage{resizegather}

\usepackage{amsmath,amsfonts,bm,amssymb,mathtools,amsthm}
\usepackage{color,xcolor}

\usepackage{hyperref}       
\hypersetup{
  colorlinks,
  linkcolor={red!50!black},
  citecolor={blue!50!black},
  urlcolor={blue!80!black}
  }
  \definecolor{orange}{HTML}{ff7f0e}
  \definecolor{blue}{HTML}{1f77b4}
  
\usepackage{url}

\usepackage{algorithm}
\usepackage{adjustbox}

\usepackage{microtype}
\usepackage{lipsum}
\usepackage{graphicx}
\usepackage{sidecap}
\usepackage{caption}
\usepackage{booktabs} 
\usepackage{amsfonts}       
\usepackage{nicefrac}       
\usepackage{microtype}      
\usepackage{comment}
\usepackage{enumitem}
\usepackage{wrapfig,tikz}
\usepackage{longtable,fancyhdr}
\usepackage{adjustbox, bigstrut, tabularx, multirow, makecell, diagbox}
\usepackage{graphicx,float}
\usepackage{stackrel}
\usepackage{color}
\usepackage{colortbl}
\usepackage{theoremref}
\usepackage{cases}
\usepackage{stmaryrd}
\usepackage{mathabx}
\usepackage{dsfont}

\usepackage[capitalise]{cleveref}

\makeatletter
\usepackage{xspace} 
\def\@onedot{\ifx\@let@token.\else.\null\fi\xspace}
\DeclareRobustCommand\onedot{\futurelet\@let@token\@onedot}

\definecolor{blue1}{RGB}{0,128,255}
\definecolor{blue3}{RGB}{0,0,128}
\definecolor{darkpastelgreen}{rgb}{0.01, 0.75, 0.24}
\definecolor{cerulean}{rgb}{0.0, 0.48, 0.65}

\newcommand*{\tran}{^{\mkern-1.5mu\mathsf{T}}}

\newcommand{\mbb}[1]{\mathbb{#1}}
\newcommand{\mcal}[1]{\mathcal{#1}}

\def\vs{\emph{vs}\onedot}

\def\iid{i.i.d\onedot}

\definecolor{darkgreen}{rgb}{0,0.6,0}

\newtheorem{theorem}{Theorem}

\newtheorem{proposition}{Proposition}

\def\eqref#1{equation~\ref{#1}}

\def\1{\bm{1}}

\def\rvv{{\mathbf{v}}}

\def\rvx{{\mathbf{x}}}
\def\rvy{{\mathbf{y}}}
\def\rvz{{\mathbf{z}}}

\def\vtheta{{\bm{\theta}}}

\def\vd{{\bm{d}}}

\def\vh{{\bm{h}}}

\def\vs{{\bm{s}}}

\def\mI{{\bm{I}}}
\def\mJ{{\bm{J}}}

\DeclareMathAlphabet{\mathsfit}{\encodingdefault}{\sfdefault}{m}{sl}
\SetMathAlphabet{\mathsfit}{bold}{\encodingdefault}{\sfdefault}{bx}{n}

\newcommand{\pdata}{p_{\rm{data}}}

\newcommand{\dt}{\frac{\partial}{\partial t}}
\newcommand{\E}{\mathbb{E}}

\newcommand{\R}{\mathbb{R}}

\DeclareMathOperator*{\argmax}{arg\,max}
\DeclareMathOperator*{\argmin}{arg\,min}

\newcommand{\ud}{\mathrm{d}}

\newcommand{\norm}[1]{\left\lVert#1\right\rVert}

\newcommand{\Dp}{{\mathcal{D}_P}}
\newcommand{\Dq}{{\mathcal{D}_Q}}
\newcommand{\bbR}{{\mathbb{R}}}
\newcommand{\bbE}{{\mathbb{E}}}
\newcommand{\bx}{{\mathbf{x}}}

\newcommand{\by}{{\mathbf{y}}}
\newcommand{\bff}{{\mathbf{f}}}
\newcommand{\bw}{{\mathbf{w}}}
\newcommand{\dd}{{\textrm{d}}}

\newcommand{\dreinf}{{\textrm{DRE-}\infty}}

\definecolor{light-gray}{gray}{0.92}  
\usepackage[accepted]{aistats2022}
\begin{document}

%

%

\twocolumn[

\aistatstitle{Density Ratio Estimation via Infinitesimal Classification}

\aistatsauthor{ Kristy Choi$^*$ \And Chenlin Meng$^*$ \And  Yang Song \And Stefano Ermon }

\aistatsaddress{ Computer Science Department, Stanford University } ]

\begin{abstract}
Density ratio estimation (DRE) is a fundamental machine learning technique for comparing two probability distributions.
However, existing methods struggle in high-dimensional settings, as 
it is difficult to accurately compare probability distributions based on finite samples.
In this work we propose $\dreinf$, a divide-and-conquer approach to reduce DRE to a series of easier subproblems.  Inspired by Monte Carlo methods, we smoothly interpolate between the two 
distributions
via an infinite continuum of intermediate bridge distributions. 
We then estimate the instantaneous rate of change 
of the bridge distributions indexed by time
(the ``time score'')---a quantity defined analogously to data (Stein) scores---with a novel time score matching objective.  
Crucially, the learned time scores can then be integrated to compute the desired density ratio. 
In addition, we show that traditional (Stein) scores can be used to obtain integration paths that connect regions of high density in both distributions, improving performance in practice. 
Empirically, we demonstrate that our approach performs well on downstream tasks such as mutual information estimation and energy-based modeling on complex, high-dimensional datasets.

\end{abstract}

\section{INTRODUCTION}
\label{intro}
\begin{figure*}[!t]
    \centering
        \includegraphics[width=0.75\textwidth]{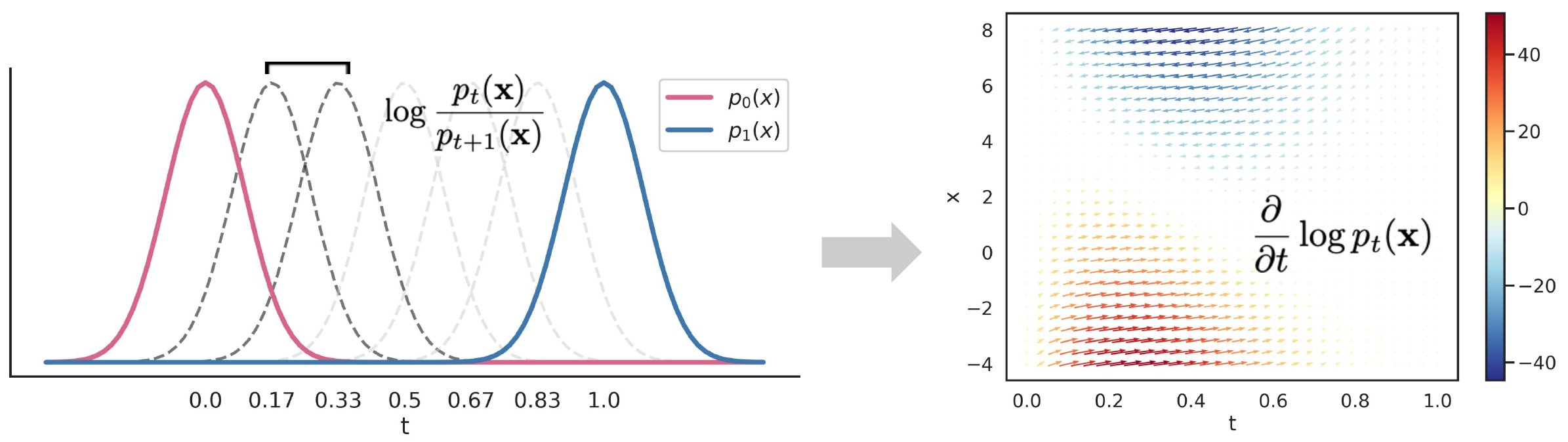}
    \caption{An overview of $\dreinf$'s time score matching framework. Instead of bridging $p_0(\rvx) \equiv q(\rvx)$ and $p_1(\rvx) \equiv p(\rvx)$ with a finite number of bridges, we smoothly interpolate and estimate the instantaneous rate of change of each intermediate distribution $\dt \log p_t(\rvx)$. The x-coordinates of the vector field on the right denote the time scores, while the y-coordinates denote data scores. Arrows are colored by the time score values.}
    \label{fig:framework}
\end{figure*}
Machine learning algorithms often require a way to compare and contrast two probability distributions
$q(\rvx)$ and $p(\rvx)$, given a set of finite samples from each distribution.
A natural quantity for such a task is the likelihood ratio $r(\rvx) = q(\rvx)/p(\rvx)$ of the two densities \citep{nguyen2007estimating,sugiyama2008direct}, which leads to the problem of \emph{density ratio estimation (DRE).}
DRE enjoys a wide range of applications such as generative modeling \citep{goodfellow2014generative,nowozin2016f}, representation learning and mutual information estimation \citep{oord2018representation,belghazi2018mutual,poole2019variational,song2019understanding}, domain adaptation \citep{gretton2009covariate,yamada2013relative}, importance sampling \citep{meng1996simulating,gelman1998simulating,neal2001annealed,sinha2020neural,yao2020adaptive}, 
and propensity score matching for causal inference \citep{abadie2016matching,shalit2017estimating,johansson2018learning}.

Despite its widespread use, accurate DRE from finite samples is challenging
in high dimensions.
A naive construction of an estimator for this likelihood ratio can require a number of samples exponential in the Kullback-Leibler (KL) divergence of the two densities to be accurate \citep{chatterjee2018sample,mcallester2020formal}.
Therefore, prior works have found success in a divide-and-conquer approach \citep{rhodes2020telescoping}.
They split the global problem into a sequence of easier DRE subproblems for $T > 0$ intermediate bridging distributions that are closer to each other, 
thereby ``shrinking'' the gap between $p(\rvx)$ and $q(\rvx)$. 
\cite{rhodes2020telescoping} takes a discriminative approach by training multiple classifiers---one for each pair of bridge
distributions---and aggregates their outputs to obtain the desired ratio estimates.
Although they demonstrate that using more intermediate distributions helps performance,
naively increasing the number of bridging distributions $T$ is undesirable.
Not only does the model size and complexity grow linearly with the number of bridges, 
but also the approach requires evaluating more classifiers at test time, which is computationally expensive. 


To address such limitations, 
we draw inspiration from 
annealed importance sampling and path sampling
\citep{neal1993probabilistic,meng1996simulating,gelman1998simulating}
to generalize this divide-and-conquer approach by considering its limiting case.
We connect $q(\rvx)$ and $p(\rvx)$ by constructing an \emph{infinite} number of bridge distributions $p_t(\rvx)$---indexed by a \emph{continuous} ``time'' variable $t \in [0,1]$---via an interpolation mechanism.
This gives our method $\dreinf$ its namesake.
The key to our approach is to estimate for each location $\rvx$ the instantaneous rate of change of the intermediate log-density $p_t(\rvx)$ along this path of bridging distributions.
The rate of change $\frac{\partial}{\partial t} \log p_t(\rvx)$ measures how each intermediate density is locally changing along a prescribed trajectory in distribution space over time.
As this is in direct analogy to the traditional (Stein) score or \emph{data score}  $(\nabla_\rvx \log p(\rvx))$, which measures how a density is changing over its input domain \citep{hyvarinen2005estimation,kingma2010regularized}, we call this quantity $\frac{\partial}{\partial t} \log p_t(\rvx)$ the \emph{time score}.
The intuition is that since 
$\frac{\partial}{\partial t} \log p_t(\rvx) \approx (\log p_{t+\Delta t}(\rvx)  - \log p_{t}(\rvx))/ \Delta t = (\log \frac{p_{t+\Delta t}(\rvx)}{p_{t}(\rvx)})/ \Delta t$,
the time score characterizes the log density ratio between two distributions with an infinitesimal gap ($\Delta t$).
%
This allows us to compute the original density ratio $r(\rvx)$ 
by \emph{integrating} (rather than summing) the time scores over $t \in [0,1]$.
Figure~\ref{fig:framework} illustrates an overview of our $\dreinf$ method.

Because the true underlying time scores are unknown, we introduce a
framework to estimate them from data.
We propose a new \emph{time score matching} objective
to efficiently train a neural network for learning the time scores.
We additionally prove that this time score matching objective is equivalent
to solving a series of ``infinitesimal classification'' tasks between two extremely close bridge distributions.
Perhaps counterintuitively, $\dreinf$
generalizes TRE to an \emph{infinite number} $T$ of bridge distributions while simultaneously overcoming its various computational limitations.
We also show that this framework also naturally allows for incorporating auxiliary information from the data scores $(\nabla_\rvx \log p_t(\rvx))$ of the bridge distributions that is helpful for estimating density ratios more accurately in practice. 
To do so, we introduce a hybrid training objective that 
jointly learns both the data and time scores,
which allows for the construction of integration paths connecting regions of high data density for both $q(\rvx)$ and $p(\rvx)$.
Empirically, we demonstrate the efficacy of our approach on downstream tasks which require access to accurate density ratios, such as mutual information estimation and energy-based modeling on complex, high-dimensional data.

In summary, the contributions of our work are:
\begin{enumerate}
    \item We propose $\dreinf$, a DRE technique that smoothly interpolates between two distributions and 
    involves learning the rate of change of the intermediate log densities (time scores).
    \item We introduce a novel framework to learn the time scores from data---time score matching---
    that allows for the scalable estimation of such time scores, and demonstrate how to leverage black-box numerical integrators to obtain density ratios.
    \item We demonstrate how to leverage data scores via a hybrid objective to improve our density ratio estimates in practice, by connecting regions of high data density in both distributions.
\end{enumerate}
\section{PRELIMINARIES}
\label{prelims}
\paragraph{Notation and Problem Setup.}
Let $p(\rvx)$ and $q(\rvx)$ be two unknown distributions over  $\mathcal{X} \in \mathbb{R}^D$,
for which we have access to \iid samples $\Dp = \{\rvx_{i}\}_{i=1}^N \sim p(\rvx)$ and $\Dq = \{\rvx_{i}\}_{i=1}^N \sim q(\rvx)$.
The goal of density ratio estimation (DRE) is to accurately estimate $r(\rvx) = q(\rvx)/p(\rvx)$ given $\Dp$ and $\Dq$.

\begin{figure*}[!t]
    \centering
        \subfigure[Baseline MSE: 952.8]{\includegraphics[width=.25\textwidth]{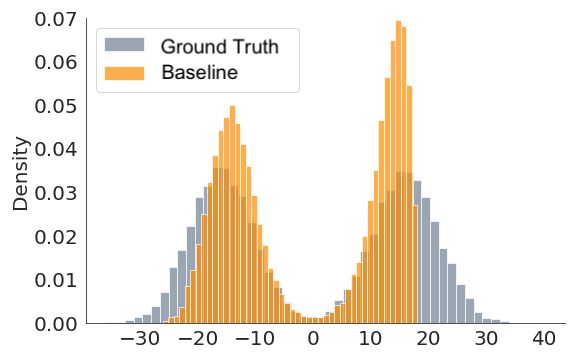}}
        \subfigure[TRE(4) MSE: 1.1]{\includegraphics[width=.25\textwidth]{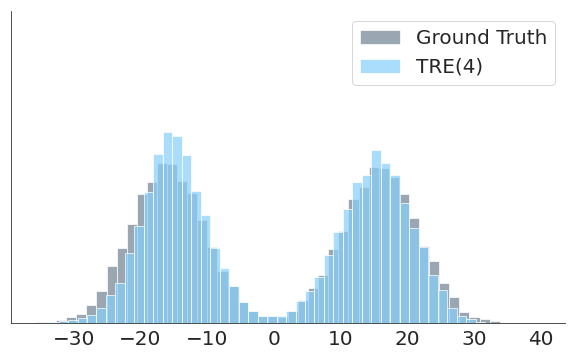}}
        \subfigure[TRE(9) MSE: 0.63]{\includegraphics[width=.24\textwidth]{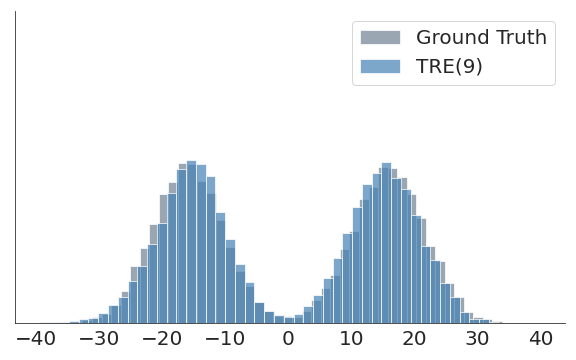}}
        \subfigure[Time MSE (Ours): 0.58]{\includegraphics[width=.24\textwidth]{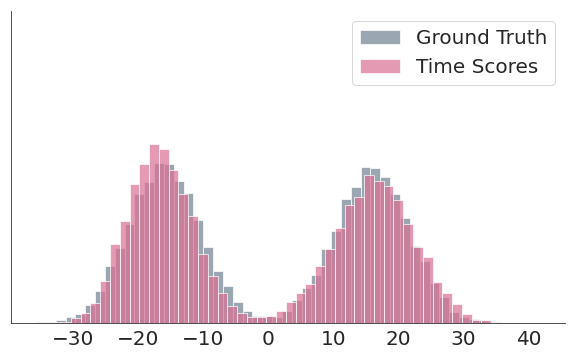}}
    \caption{Motivating example on a synthetic 2-D Gaussian dataset, with learned density ratio estimates by method relative to the ground truth values for (a-d). The performance of TRE improves with more intermediate bridge distributions, while DRE-$\infty$ outperforms the rest. 
    The x-axis denotes the log-ratios.}
    \label{fig:2d_exp}
\end{figure*}

\paragraph{DRE via Probabilistic Classification.}
\label{bin_clf}
A well-known technique for DRE is probabilistic classification, where a binary classifier $h_\theta: \mathcal{X} \rightarrow [0,1]$ is trained to discriminate between two sets of samples $\Dp$ and $\Dq$ \citep{sugiyama2008direct,menon2016linking}. 
Each ``dataset'' from a particular distribution is assigned a pseudolabel of either $y=0$ or $y=1$ depending on its source.
Once the classifier has been trained, the corresponding ratios can be recovered from its class 
probabilities via Bayes Rule:
\begin{equation}
\label{naive}
    r(\rvx) = \frac{q(\rvx)}{p(\rvx)} = \frac{p(\rvx \mid y=1)}{p(\rvx \mid y=0)} =  \frac{h_\theta^*(\rvx)}{1-h_\theta^*(\rvx)} 
\end{equation}
where $h_\theta^*$ denotes the Bayes optimal classifier.
Despite its elegant simplicity, this method fails in settings where $q(\rvx)$ and $p(\rvx)$ are sufficiently different.
The discriminative task
becomes trivial, and the classifier can achieve perfect accuracy but fail to estimate the ratios correctly due to poorly calibrated output probabilities \citep{rhodes2020telescoping,choi2021featurized}.
Such a failure mode is illustrated on a simple DRE task for 2-dimensional Gaussians in Figure~\ref{fig:2d_exp}(a), where $p(\rvx) = \mcal{N}(\bm{0}, \mI)$ and $q(\rvx) = \mcal{N}(\bm{4}, \mI)$.
In particular, the classifier 
cannot capture the entire range of variation of the likelihood ratios (see \cref{fig:2d_exp}(a)).

\paragraph{Improving Estimates with Bridges.}
To sidestep this challenge, 
Telescoping Density Ratio Estimation (TRE) \citep{rhodes2020telescoping} proposes a \emph{divide-and-conquer} approach 
and partitions the binary classification problem in \cref{naive} into several easier subproblems.
TRE constructs $T > 0$ bridge densities $\{p_t(\rvx)\}_{t=1}^{T}$ by interpolating between $p_0(\rvx) \equiv q(\rvx)$ and $p_1(\rvx) \equiv p(\rvx)$, and trains $T$ (conditional) classifiers to discriminate between samples from $p_t(\rvx)$ and $p_{t+1}(\rvx)$. 
Intuitively, each of the likelihood ratios $\log r_t(\rvx) = \log \frac{p_t(\rvx)}{p_{t+1}(\rvx)}$ are better behaved and thus easier to learn than those in the original problem, as in Figure~\ref{fig:2d_exp}(a).
Then,
a telescoping sum of the intermediate classifier outputs allows for the recovery of the desired likelihood ratio:
\begin{equation}
\label{eq:tre}
\begin{split}
          \log r(\rvx) &= \log q(\rvx) - \log p(\rvx) = \sum_{t=1}^{T} \log r_t(\rvx) \\
          &= \sum_{t=1}^{T} \log p_{t}(\rvx) - \log p_{t+1}(\rvx) 
\end{split}
\end{equation}

Figures~\ref{fig:2d_exp}(b)-(c) demonstrate that TRE with 4 and 9 intermediate distributions respectively dramatically improves performance for our 2-D task, as the histograms of the learned ratios significantly overlap with those of ground truth.
However, naively increasing $T$ is undesirable for a number of reasons.
Not only does the model 
size and complexity in TRE grow with $T$, but also the approach requires the evaluation of multiple classifiers at test time, which can be prohibitively expensive. 

\section{FROM DISCRETE BRIDGES TO CONTINUOUS PATHS}
\label{method}
As an alternative to \cref{eq:tre}, we consider a continuous \textit{path} of $T \rightarrow \infty$ bridge distributions connecting $p_0(\rvx) \equiv q(\rvx)$ and $p_1(\rvx) \equiv p(\rvx)$ in distribution space. 
Concretely, we denote this sequence of probability densities indexed by $t \in [0, 1]$ as $\{p_{t}(\rvx)\}_{t=0}^1$ and let $p(t) = \mathcal{U}[0,1]$ denote a uniform distribution over time steps.
We note that there are several ways to construct the intermediate bridges $p_t(\rvx)$.
A benefit of DRE-$\infty$ (which also holds for TRE) is that
it only requires we be able to efficiently \emph{sample} from $p_t(\rvx)$ without necessarily knowing its analytical form.
For example, we can define 
$\rvx(t) := \sqrt{1-\alpha(t)^2} \rvx + \alpha(t) \rvy$, where $\rvx \sim q$, $\rvy \sim p$, $\rvx(t) \sim p_t$, and $\alpha: [0,1] \to \R^+$ is a positive function that satisfies $\alpha(0) = 0$ and $\alpha(1) = 1$. 


To build some intuition for $\dreinf$, we observe the behavior of the log density ratio between two bridge distributions as the gap ($\Delta t = \frac{1}{T}$) between $p_t(\rvx)$ and $p_{t+1}(\rvx)$ becomes infinitesimal. 
Using finite differences, we can see that the intermediate densities $p_t(\rvx)$ are changing at each timestep $t$ by: $(\log p_{t+\Delta t}(\rvx)  - \log p_{t}(\rvx))/ \Delta t = \left(\log \frac{p_{t+\Delta t}(\rvx)}{p_{t}(\rvx)}\right)/ \Delta t \approx \dt \log p_t(\rvx)$. 
As $T \rightarrow \infty$ (and therefore $\Delta t \rightarrow 0$), this demonstrates that the object of interest is now not the individual log-ratios $\log r_t(\rvx)$ as in \cref{eq:tre}, but the instantaneous rate of change of the intermediate log densities $\dt \log p_t(\rvx)$, which we denote as the \emph{time score}. 

We formalize this intuition in the following proposition. 
The identity is well known in the path sampling literature and we include it here for completeness \citep{gelman1998simulating,owen2013monte,yao2020adaptive}.

\begin{proposition}
\label{sum_to_integral} Let $\log r(\rvx)$ denote the log density ratio between the two densities $p_0(\rvx)$ and $p_1(\rvx)$.
When $T \rightarrow \infty$, we have the following:
\begin{equation}
\resizebox{\linewidth}{!}{$
    \begin{split}
    \log r(\rvx) = \log \frac{p_0(\rvx)}{p_1(\rvx)} &= \sum_{t=1}^{T} \log \left( \frac{p_{(t-1)/T}(\rvx)}{p_{t/T}(\rvx)} \right)
    = \int_{1}^0 \frac{\partial}{\partial \lambda} \log p_\lambda(\rvx) d\lambda 
    \end{split}
$}.
\end{equation}
\end{proposition}
We provide a more detailed derivation in Appendix~\ref{path_sampling}.

There are two key takeaways from \cref{sum_to_integral}. 
First, 
as the number of bridge distributions increase to infinity, the telescoping sum in \cref{eq:tre} becomes an integral. 
The integration is important because it can be computed using any off-the-shelf numerical integrator---the fact that it is one-dimensional also means that it will be very efficient to compute.
This eliminates the need to evaluate all $T$ intermediate classifiers at inference time as in \cref{eq:tre}. 
Instead, we have an additional degree of freedom where we can choose how accurately we want to estimate $\int_{1}^0 \dt \log p_t(\rvx) dt$ by specifying the error tolerance of the numerical integrator.
This will adaptively determine the total number of necessary function (intermediate classifier) evaluations, making $\dreinf$'s inference procedure more efficient than that of TRE.
In fact, a surprising insight of $\dreinf$ is that taking the number of bridges to the infinite limit actually confers significant computational benefits over the finite regime of TRE. 

Next, we see that the log ratios $\log r_t(\rvx)$ in \cref{eq:tre} become the \emph{time scores} $\dt \log p_t(\rvx)$ in \cref{sum_to_integral}.
The time score captures---for each input $\rvx$---the instantaneous rate of change of the intermediate $p_t(\rvx)$ along the prescribed path in distribution space.
But as we rarely have access to the true time scores in most practical scenarios, we must \emph{learn} them from data.

\section{TIME SCORE MATCHING VIA INFINITESIMAL CLASSIFICATION}
\label{time_training}
We propose to train a time score model $s_\vtheta^{\textrm{time}}(\rvx, t)$ to estimate the true time scores $\dt \log p_t(\rvx)$ via the following objective:
\begin{equation}
\label{time_obj_first}
\resizebox{\linewidth}{!}{$
\begin{split}
        \mathcal{J_{\text{time}}(\theta)} &= \E_{p(t)}\E_{p_t(\rvx)}\bigg[\lambda(t) \left( \dt \log p_t(\rvx) - s_\vtheta^{\textrm{time}}(\rvx, t) \right)^2 \bigg]
\end{split}
$}
\end{equation}
where $\lambda(t): [0,1] \rightarrow \bbR_+$ is a positive weighting function and $p(t)$ is the uniform distribution over timescales.
While this objective is clearly minimized when $s_\vtheta^{\textrm{time}}(\rvx, t) = \dt \log p_t(\rvx)$, it might not seem possible to evaluate and optimize it since $\dt \log p_t(\rvx)$ is unknown. 
However, using integration by parts as in \citep{hyvarinen2005estimation}, 
this objective
can be simplified to the following expression in \cref{prop:time_objective}.
As used in traditional score matching techniques \citep{kingma2010regularized,song2020sliced,song2019generative}, integration by parts allows us to obtain a practical objective function for training  $s_\vtheta^{\textrm{time}}(\rvx, t)$ that does not depend on the true time scores $\dt \log p_t(\rvx)$. 

\begin{proposition}[Informal]
\label{prop:time_objective}
Under certain regularity conditions, the optimal solution $\theta^{*}$ of \cref{time_obj_first} is the same as the optimal solution of:
\begin{equation}
\resizebox{\linewidth}{!}{$
\begin{split}
        &\mathcal{L_{\text{time}}(\theta)} = 2 \E_{q(\rvx)}[\lambda(0) s_\vtheta^{\textrm{time}}(\rvx, 0)] - 2\E_{p(\rvx)}[\lambda(1) s_\vtheta^{\textrm{time}}(\rvx, 1)] \\
    &+\E_{p(t)}\E_{p_t(\rvx)}\bigg[2 \lambda(t) \dt s_\vtheta^{\textrm{time}}(\rvx, t) + 2\lambda'(t) s_\vtheta^{\textrm{time}}(\rvx, t) + \lambda(t) s_\vtheta^{\textrm{time}}(\rvx, t)\bigg]
\end{split}
$}
\end{equation}
\end{proposition}
where the first two terms of \cref{prop:time_objective} denote the boundary conditions for $t=\{0,1\}$, and the expectations can be approximated via Monte Carlo. 
Our ability to estimate the objective via Monte Carlo samples makes training $\dreinf$
much more efficient than TRE, which requires the explicit construction of $T$ additional batches per gradient step for all intermediate classifiers.
Even with a fixed batch size $B$ that is divided among $T$ intermediate classifiers, TRE requires that $B$ grow with $T$ (since it is necessary that $B/T \geq 1$ to train each classifier).
We defer the exact assumptions and proof to Appendix~\ref{pf-time-prop}, and provide pseudocode in Appendix~\ref{app-pseudo}.

The optimal time score model, denoted by $s_{\vtheta^\ast}^{\textrm{time}}(\rvx, t)$, satisfies $s_{\vtheta^*}^{\textrm{time}}(\rvx, t) \approx \dt \log p_t(\rvx)$. Therefore after training, the log-density-ratio can be estimated by:
\begin{equation}
    \label{eqn:dr1}
    \log r(\rvx) \approx \int_{1}^0 s_{\vtheta^*}^{\textrm{time}}(\rvx, t) \ud t. 
\end{equation}

\subsection{Joint Training Objective}
We can also incorporate
helpful auxiliary information from the data scores
into the training objective in \cref{time_obj_first}.
Specifically, we can define a vector-valued score model $\vs_\vtheta^{\textrm{joint}}(\rvx, t)$ and train it with a hybrid objective that seeks to \emph{jointly learn} $\nabla_{[\rvx; t]} \log p(\rvx, t)$:
\begin{equation}
\label{eqn:joint_obj}
\resizebox{\linewidth}{!}{$
        \mathcal{J_{\text{joint}}(\theta)} =
        \E_{p(t)}\E_{p_{t}(\rvx)}\left[ \frac{1}{2}\lambda(t)\norm{\nabla_{[\rvx; t]}\log p(\rvx, t) - \vs_\vtheta^{\textrm{joint}}(\rvx, t)}_2^2 \right].
$}
\end{equation}
The data score component in \cref{eqn:joint_obj} can be obtained by DSM \citep{vincent2011connection} or SSM \citep{song2020sliced}.
For the SSM variant, \citep{hyvarinen2005estimation,song2020sliced} shows that optimizing \cref{eqn:joint_obj} is equivalent to optimizing the  following objective.
\begin{theorem}[Informal]
\label{thm:objective}
Under certain regularity conditions, the solution to the optimization problem in \cref{eqn:joint_obj} can be written as follows:
\begin{equation}
\label{eq:objective_simplified}
\resizebox{0.9\hsize}{!}{$
\begin{split}
    &\theta^{*} = \argmin_{\theta} \textrm{}
    \E_{p(t)}\E_{p_{t}(\rvx)}\E_{p({\rvv})}\bigg[ \\
    &\frac{1}{2}\lambda(t) \norm{\vs_\vtheta^{\textrm{joint}}(\rvx, t)[\rvx]}_2^2 + \lambda(t) \rvv\tran \nabla_{\rvx} \vs_\vtheta^{\textrm{joint}}(\rvx, t)[\rvx] \rvv \\
    &+ \lambda(t) \frac{\partial}{\partial t}\vs_\vtheta^{\textrm{joint}}(\rvx, t)[t] + \lambda'(t) \vs_\vtheta^{\textrm{joint}}(\rvx, t)[t]\bigg] \\
    &+ \E_{p_0(\rvx)}[\lambda(0) \vs_\vtheta^{\textrm{joint}}(\rvx, 0)[t]] - \E_{p_1(\rvx)}[\lambda(1) \vs_\vtheta^{\textrm{joint}}(\rvx, 1)[t]].
\end{split}
$}
\end{equation}
\end{theorem}
where $\rvv \sim p(\rvv) = \mathcal{N}(\bm{0}, \bm{I})$, $\vs_\vtheta^{\textrm{joint}}(\rvx, t)[\rvx]$ denotes the data score component of $\vs_\vtheta^{\textrm{joint}}$, and $\vs_\vtheta^{\textrm{joint}}(\rvx, t)[t]$ denotes the time-score component of $\vs_\vtheta^{\textrm{joint}}$.
We defer the proof and detailed assumptions to Appendix~\ref{app-joint-proof}. In practice, the expectation in \cref{eq:objective_simplified} can be approximated via Monte Carlo sampling.
We can leverage DSM when
the data follows a known stochastic differential equation (SDE) and $p(\bx) = \mathcal{N}(\bm{0}, \bm{I})$
since the analytical form of $p_{t}(\bx)$ is tractable. 

\subsection{Link to Infinitesimal Classification}
Recall that our motivation was to generalize TRE by taking the number of intermediate bridge distributions $T$ to the infinite limit.
With an infinite number of bridges, each intermediate classifier is tasked with distinguishing samples from two bridge distributions $p_{t+\Delta t}(\rvx)$ and $p_{t}(\rvx)$.
In fact, the following proposition states that the optimal form of this infinitesimal classifier 
is given by the time score $\dt \log p_t(\rvx)$.
\begin{proposition}
\label{prop:score_clf} When $T \rightarrow \infty$, the Bayes-optimal
classifier between two adjacent bridge distributions $p_{t/T}(\rvx)$ and $p_{(t+1)/T}(\rvx)$ for any $t \in [0,1]$ is: 
\begin{equation}
    \vh_{\vtheta^*}(\rvx, t) = \frac{1}{2} + \frac{1}{4} \left(\dt \log p_t(\rvx)\right) \Delta t + o(\Delta t).
\end{equation}
\end{proposition}
where $\Delta t = \frac{1}{T}$, 
and $\vh_{\vtheta^*}(\rvx, t) \in [0,1]$ is a conditional 
probabilistic classifier. 

While the above result is instructive, it does not provide us with a practical algorithm for time score estimation---
we cannot train an infinite number of such binary classifiers.
To tackle this challenge,
we consider the limit of the binary cross-entropy loss function for the optimal infinitesimal classifier (\cref{prop:score_clf}) when $T\to \infty$.
\begin{proposition}
\label{prop:opt_clf}
Let $\Delta t = 1/T$ and parameterize the binary classifier as $\vh_\vtheta(\rvx, t) = \frac{1}{2} + \frac{1}{4} \vs_\vtheta^{\textrm{time}}(\rvx, t)\Delta t$, where $\vs_\vtheta^{\textrm{time}}(\rvx, t) \approx \dt \log p_t(\rvx)$ denotes a time score model. Then from the binary cross-entropy objective:
\begin{equation}
\resizebox{\linewidth}{!}{$
\begin{split}
    &\arg \max_{\theta} \mathbb{E}_{p_{t}(\rvx)}[\log (1 - \vh_\vtheta(\rvx, t))] + \mathbb{E}_{p_{t+\Delta t}(\rvx)}[\log \vh_\vtheta(\rvx, t)]\\
    =& \arg \max_{\theta} -\frac{1}{4} (\Delta t)^2 \mbb{E}_{p_t(\rvx)}\left[\left(\vs_\vtheta^{\textrm{time}}(\rvx, t) - \dt \log p_t(\rvx) \right)^2\right] + o((\Delta t)^2)
\end{split}
$}
\end{equation}
\end{proposition}
We defer the proof to Appendix~\ref{pf-opt-clf}. 
Notably, the form of the objective function in \cref{prop:opt_clf} exactly mirrors that of \cref{time_obj_first}, drawing the equivalence between solving an infinite number of ``infinitesimal'' classification problems and time score matching. 
This extends the previous connection between infinitesimal classification problems and score matching as mentioned in \citep{gutmann2012bregman} and \citep{ceylan2018conditional} in the context of estimating unnormalized probability models.
\section{LEARNING TIME SCORES IN PRACTICE}
\subsection{Variance Reduction via Importance Weighting} 
\label{impl-design}
In our preliminary experiments, we found that a naive implementation of \cref{prop:time_objective} led to unstable training due to the high variance in the objective across the different timescales $t$.
This finding is in accordance with recent work on diffusion probabilistic models \citep{nichol2021improved,kingma2021variational,song2021maximum}, which emphasize the critical role of applying the proper weighting function $\lambda(t)$ rather than randomly sampling $t \sim \mathcal{U}[0,1]$.
This problem is also exacerbated by the fact that our training objective requires backpropagating through the score network $s_\vtheta^{\textrm{
time}}(\rvx, t)$ with respect to $t$, which can cause training to diverge or progress extremely slowly for certain design choices.

Drawing inspiration from \citet{nichol2021improved,song2021maximum,kingma2021variational}, we learn an importance weighting scheme of the distribution over timescales $p(t)$.
We optimize the following  importance-weighted objective rather than \cref{prop:time_objective}:
\begin{equation}
\label{rw_time_obj}
\resizebox{\linewidth}{!}{$
\begin{split}
        \mathcal{J_{\text{rw-time}}(\theta)} &= \E_{p_{\text{iw}}(t)}\E_{p_t(\rvx)}\bigg[\frac{p(t)\lambda(t)}{p_{\text{iw}}(t)} \left( \dt \log p_t(\rvx) - s_\vtheta^{\textrm{
time}}(\rvx, t) \right)^2 \bigg]
\end{split}
$}
\end{equation}
where $p(t) \sim \mcal{U}[0,1]$ and $p_{\text{iw}}(t)$ is a \emph{learned} proposal distribution.
We approximate the importance weighting distribution by maintaining a history buffer of the $B$ most recent loss values in \cref{prop:time_objective} (excluding the boundary conditions which are constant w.r.t. $t$).
Then, we use this buffer to estimate an importance sampling distribution $p_{\text{iw}}(t)$ over $t$ designed to reduce the variance of the loss. We report specific implementation details in Appendix~\ref{app:loss_history}.

\subsection{Incorporating Auxiliary Information via Data Scores}
\label{joint_training}

Another advantage of $\dreinf$
is that it allows for considerable flexibility in the way that the likelihood ratios $r(\rvx)$ are computed.
Recall that 
in \cref{eqn:dr1}, the time scores $\dt \log p_t(\rvx)$ are integrated over $t \in [0,1]$ while $\rvx$ is fixed along a horizontal path.
However, we are not required to stick to this simple integral.
We can actually construct \emph{arbitrary} paths---that is, vary both $\rvx$ and $t$ in the integral---with a theoretical guarantee that we will recover the same density ratios as before.
We find that this approach often helps improve performance in practice, and call it the ``pathwise method."

\begin{figure}[!t]
    \centering
        \subfigure{\includegraphics[width=.5\textwidth]{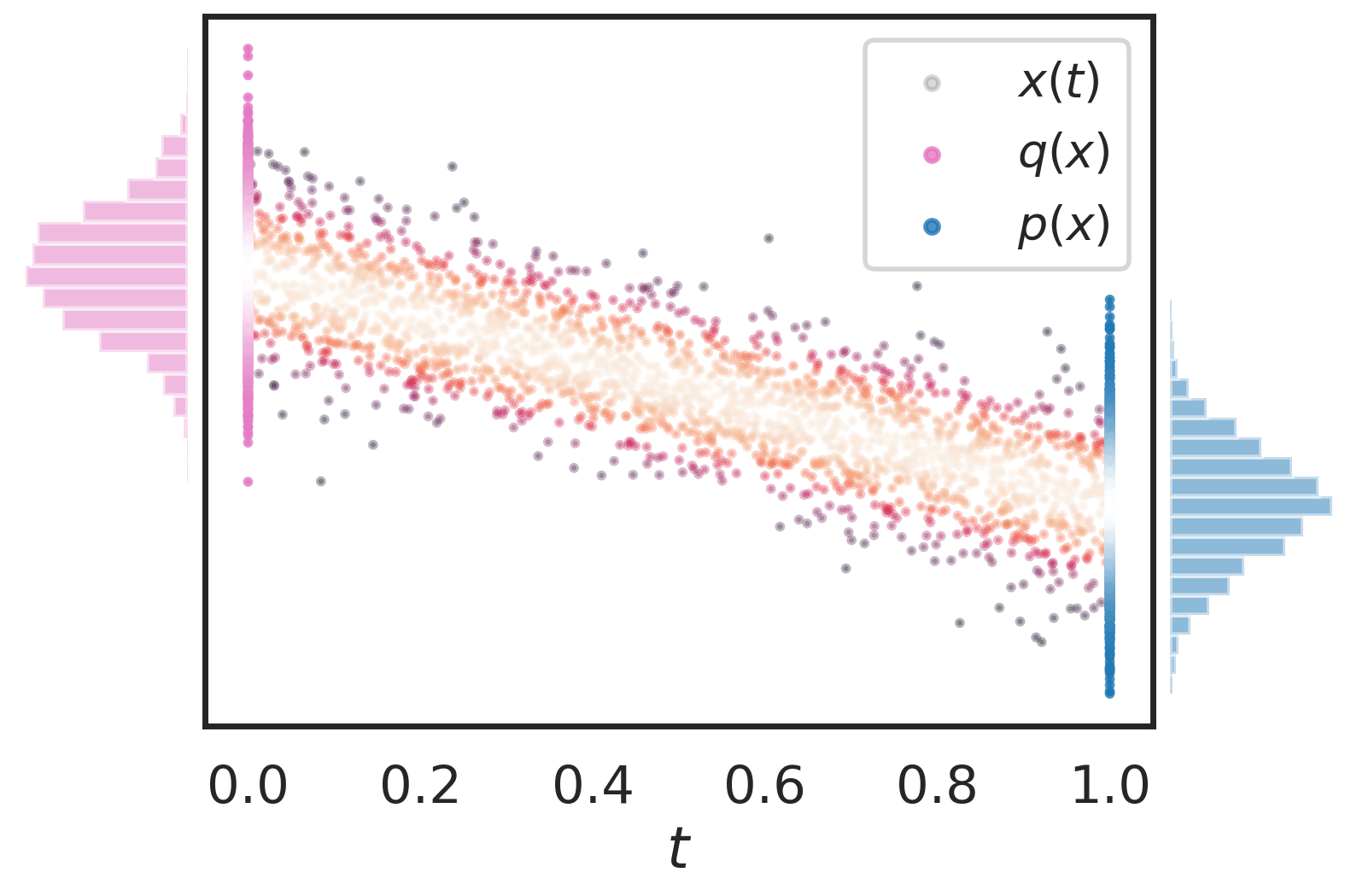}}
    \caption{
    An example of the simple line path $\rvy(t) = \rvx + t \cdot (\rvz - \rvx)$ bridging the high-density regions of $q(\rvx)$ and $p(\rvx)$, where $\rvz$ is sampled from $p(\rvx)$. Brighter color indicates higher density.}
    \label{fig:bridge_err_simple_paths}
\end{figure}

The pathwise method aims to compute a variant of \cref{eqn:dr1} by evaluating our time score model at various points $\rvy(t)$ where its estimates will be the most accurate.
To do so, we prescribe a path $\rvy(t)$ such that it connects 
$\rvx$ in the high data density region of $p_0(\rvx) \equiv q(\rvx)$ to $\rvz$ in the high data density region of $p_1(\rvx) \equiv p(\rvx)$.
This trajectory can be described by an ordinary differential equation (ODE):
\begin{align*}
    \begin{cases}
    \rvy'(t) = \vd(\rvy, t)\\
    \rvy(0) = \rvx.
    \end{cases}
\end{align*}
where $\vd$ is any function that captures the relationship between $\rvy$ and $t$. 
A simple example of such a path is the line $\rvy(t) = \rvx + t \cdot (\rvz - \rvx)$ as shown in Figure~\ref{fig:bridge_err_simple_paths}, where $\rvy'(t) = (\rvz - \rvx)$. 
A reasonable choice for $\rvz$ in this case are samples from $p(\rvx)$, but we note that there are several possible choices for the path connecting $p(\rvx)$ and $q(\rvx)$ \citep{gelman1998simulating}. 

Using this path, the difference between $\log q(\rvx)$ and $\log p(\rvy)$ can be obtained by integration:
\begin{align*}
    &\log q(\rvx) - \log p(\rvy) \\
    &= \int_1^0 \dt \log p_t(\rvy(t)) + \vd(\rvy(t), t)\tran \nabla_\rvx \log p_t(\rvy(t)) \ud t.
\end{align*}
Finally, we can compute the density ratio via the following expression:
\begin{equation}
    \begin{split}
    \label{eq:pathwise}
    &r(\rvx) = (\log q(\rvx) - \log p(\rvy)) + (\log p(\rvy) - \log p(\rvx))\\
    &= \underbrace{\bigg(\int_1^0 \dt \log p_t(\rvy(t)) + \vd(\rvy(t), t)\tran \nabla_\rvx \log p_t(\rvy(t)) \ud t \bigg)}_{\text{Term 1}} \\
    &+ \underbrace{\bigg(\int_1^0 \nabla_\rvx \log p(\rvy + t(\rvx - \rvy))\tran (\rvx - \rvy) \ud t\bigg)}_{\text{Term 2}}.
    \end{split}
\end{equation}
\cref{eq:pathwise} decomposes the density ratio into 2 terms, where the first term depends on both the time score and data score, while the second term only depends on the data score.
The integrals in \cref{eq:pathwise} can be approximated using off-the-shelf ODE solvers.
Note that when the density of $p$ is tractable (e.g. a Gaussian distribution, as in energy based modeling), the second term $\log p(\rvy) - \log p(\rvx)$ can be computed in closed form. 
This property of the pathwise method 
makes it a particularly attractive alternative to \cref{eqn:dr1}.
We note that this method can be trained with the joint objective function in \cref{eqn:joint_obj}.

\begin{figure*}[!t]
    \centering
        \subfigure[TRE(4) MSE: 2.8]{\includegraphics[width=.25\textwidth]{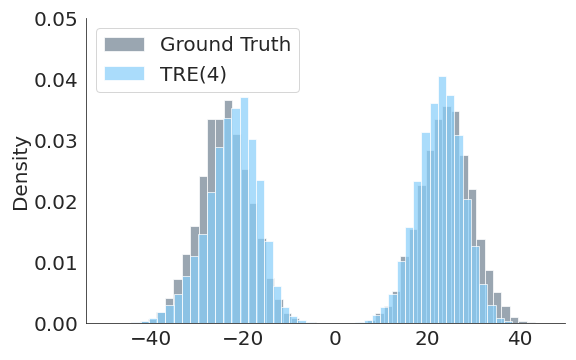}}
        \subfigure[TRE(9) MSE: 1.8]{\includegraphics[width=.24\textwidth]{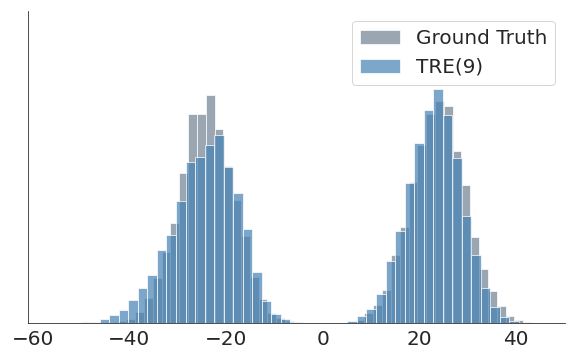}}
        \subfigure[Time MSE (Ours): 1.0]{\includegraphics[width=.24\textwidth]{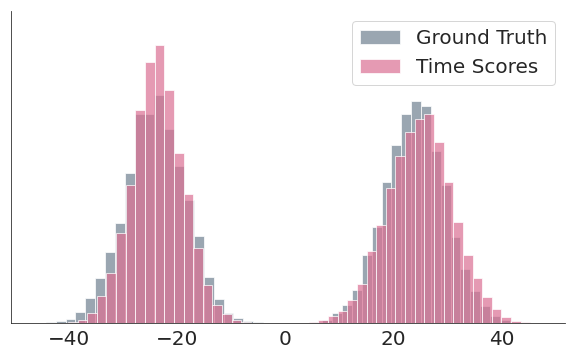}}
        \subfigure[Joint MSE (Ours): 0.6]{\includegraphics[width=.25\textwidth]{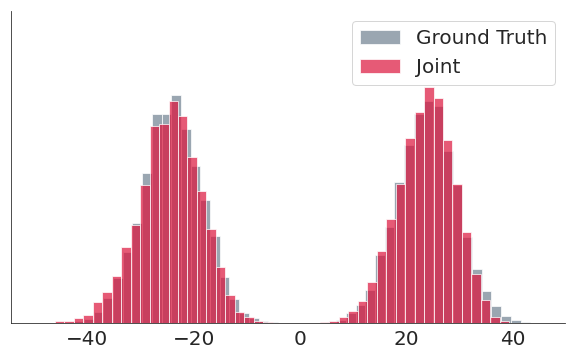}}
    \caption{Additional results on the synthetic 2-D Gaussian dataset on a more challenging evaluation task, where half the samples are shifted by 1. While all models' performance slightly degrade, our joint score matching objective still accurately recovers the density ratio estimates. The x-axis denotes the log ratios.}
    \label{fig:2d_exp_shift1}
\end{figure*}

\section{EXPERIMENTAL RESULTS}
\label{experiments}
In this section, we are interested in empirically investigating the following questions:
\begin{enumerate}
    \item Does $\dreinf$ lead to more accurate density ratio estimation than existing baselines?
    \item Does incorporating auxiliary information for DRE (e.g. learning the data scores) help to learn more accurate ratios?
\end{enumerate}

\subsection{Synthetic Gaussian Experiments}
\label{sec:synthetic}
Our running example with the 2-D synthetic dataset of Gaussian mixtures is comprised of 10K samples each from $p(\rvx) = \mathcal{N}(\bm{0}, \bm{I})$ and $q(\rvx) = \mathcal{N}(\bm{4}, \bm{I})$.
We use the Variance Preserving SDE (VPSDE) noise schedule as in \citep{ho2020denoising,song2020score} for the construction of $\rvx(t)$.
Thus $\rvx(t) \sim q(\rvx)$ when $t=0$ and $\rvx(t) \sim p(\rvx)$ when $t=1$.
Our score networks are fully-connected MLPs with ELU activation functions, with the time conditioning signal concatenated to the inputs before feeding them into the network.
As shown in Figure~\ref{fig:2d_exp}, 
our $\dreinf$ outperforms all baselines with a finite number (0, 4, 9) of intermediate distributions.

Additionally, the benefits of the pathwise approach (Section~\ref{joint_training}) is shown in Figure~\ref{fig:2d_exp_shift1}(d), where all models trained on $p(\rvx) = \mcal{N}(\bm{0}, \mI)$ and $q(\rvx) = \mcal{N}(\bm{4}, \mI)$ are evaluated on 10K samples drawn from $\mcal{N}(\bm{0}, \mI)$ and $\mcal{N}(\bm{5}, \mI)$ each.
Although all models perform worse than in Figure~\ref{fig:2d_exp} due to the slight mismatch in train and test conditions (the baseline binary classifier is not shown, as it performed extremely poorly), the additional integration path helps the jointly trained score model to more accurately recover the density ratios relative to other methods. 
We also note that the pathwise approach yielded the lowest MSE of 0.35 among all methods shown in Figure~\ref{fig:2d_exp}.
We refer the reader to Appendix~\ref{app-arch} for more details on the model architecture and hyperparameter settings, as well as Appendix~\ref{addtl-synthetic-app} for additional synthetic experiments on 1-D problems.


\subsection{Mutual Information Estimation for High-Dimensional Gaussians}
\label{sec:mi}
\begin{figure}[!ht]
    \centering
        \includegraphics[width=\linewidth]{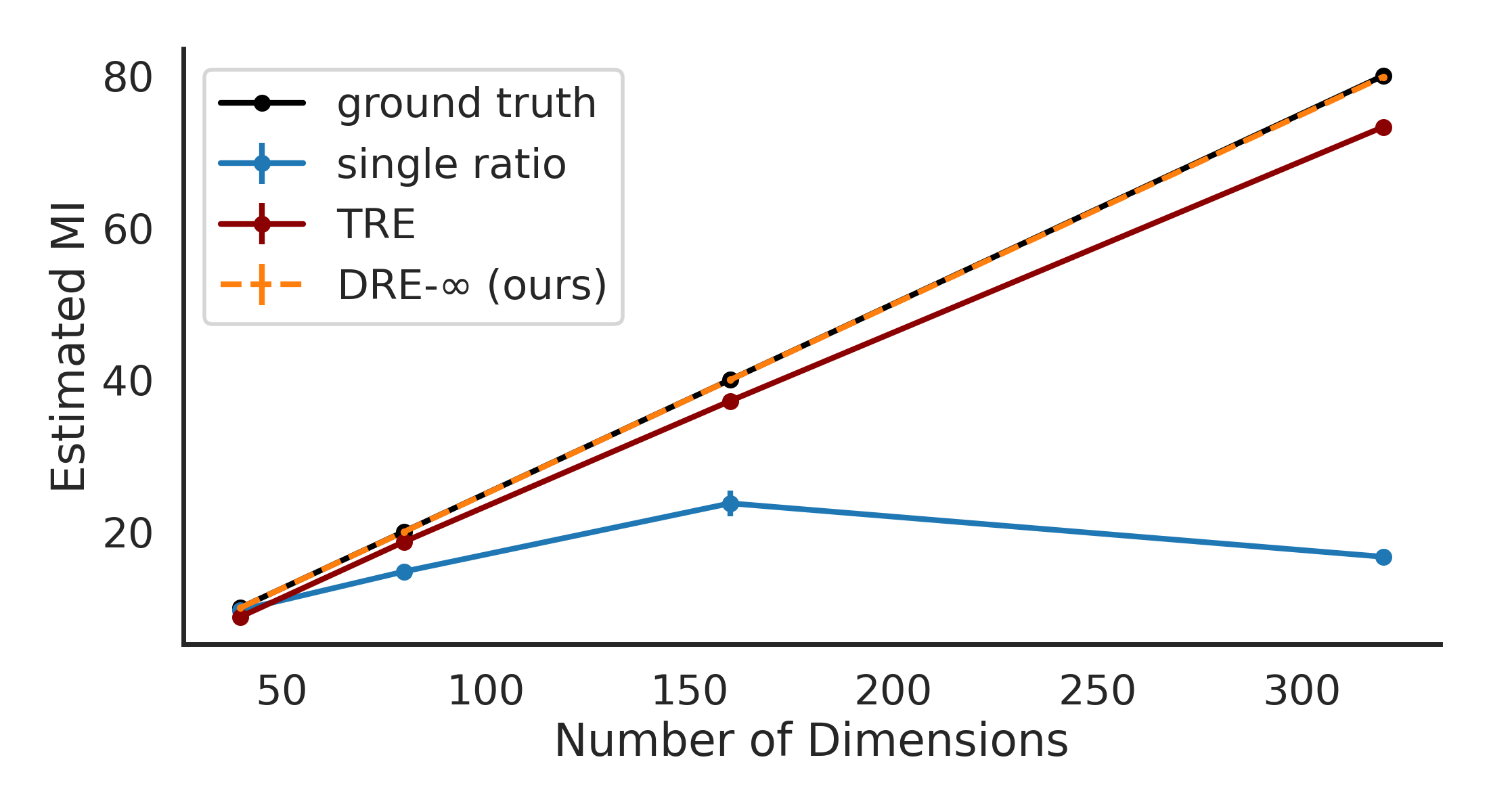}
    \caption{Estimated MI between two correlated high-dimensional Gaussian random variables, where our joint score matching objective outperforms TRE in all settings. Results are averaged over 3 runs.}
    \label{fig:mi}
\end{figure}

Next, we evaluate our approach on a mutual information (MI) estimation task between two correlated, high-dimensional Gaussians. MI estimation between two random variables is a direct application of DRE, as the problem can be reduced to estimating average density ratios between their joint density and the product of their marginals: $I(\rvx;\rvy) = \mathbb{E}_{p(\bx,\by)}\left[\log \frac{p(\bx,\by)}{q(\bx)p(\by)}\right]$. We adapt the experimental setting of \citep{belghazi2018mutual,poole2019variational,rhodes2020telescoping}, where we sweep over the dimensions $d=\{40,80,160,320\}$, and fix the correlation coefficient to be $\rho=0.8$. 

For the TRE baseline, we use the default hyperparameter settings in \citep{rhodes2020telescoping}. 
We use the joint score matching objective for our method, where we use the 
same interpolation procedure as in the 2-D Gaussian experiment in Section~\ref{sec:synthetic}.
We find that our method's estimated MI values overlap with the ground truth in all settings as shown in Figure~\ref{fig:mi}.
DRE-$\infty$ outperforms TRE in all cases and the performance gap between the two methods increase in higher dimensions.
A single binary classifier, on the other hand, fails completely for dimensions greater than $d=40$.
For additional details on the experimental setup, we refer the reader to Appendix~\ref{appendix_mi}. 

\begin{table*}[!t]
\centering
{
\setlength{\extrarowheight}{1.5pt}
\begin{adjustbox}{max width=\linewidth}
\begin{tabular}{cc|c|ccc}
  \Xhline{3\arrayrulewidth}
  \textbf{Method} & \textbf{Interpolation} & Noise & Direct ($\downarrow$) & RAISE ($\downarrow$) & AIS ($\downarrow$) \\
  \Xhline{\arrayrulewidth}
  NCE & Gaussian & 2.01 & 1.96 &  1.99 & 2.01 \\
  TRE & Gaussian & 2.01 & 1.39 & 1.35 & 1.35 \\
  \rowcolor{light-gray}
  \textbf{$\dreinf$} & Gaussian & 2.01 & \textbf{1.33} & \textbf{1.33} &  \textbf{1.33} \\
  \Xhline{\arrayrulewidth}
  NCE & Copula & 1.40 & 1.33 &  1.48 &  1.45 \\
  TRE & Copula & 1.40 & 1.24 & 1.23 & 1.22 \\
  \rowcolor{light-gray}
  \textbf{$\dreinf$} & Copula & 1.40 & \textbf{1.21} & \textbf{1.21} & \textbf{1.21} \\
  \Xhline{\arrayrulewidth}
  NCE & RQ-NSF & 1.12 & 1.09 &  1.10 &  1.10 \\
  TRE & RQ-NSF & 1.12 & 1.09 & 1.09 & 1.09 \\
  \rowcolor{light-gray}
  \textbf{$\dreinf$} & RQ-NSF & 1.12 & 1.09 & \textbf{1.08} & \textbf{1.08} \\
  \Xhline{3\arrayrulewidth}
\end{tabular}
\end{adjustbox}
}
\caption{Estimated log-likelihood results on the energy-based modeling task for MNIST, reported in bits per dimension (bpd). Lower is better. Results for NCE and TRE are from \citep{rhodes2020telescoping}. We note that $\dreinf$'s time score matching framework leads to performance improvements over relevant baselines in all settings.
}
\label{table:mnist_ebm}
\end{table*}

\subsection{Energy-based Modeling with MNIST}
\label{sec:mnist}
In this experiment, we train an energy-based model (EBM) of the MNIST dataset \citep{lecun1998mnist} using time-wise score matching. 
Specifically, we let $q(\bx)$ denote the distribution over MNIST digits and experiment with three different settings for $p(\rvx)$ as in \citep{rhodes2020telescoping}: a Gaussian noise model, a Gaussian copula, and a Rational Quadratic Neural Spline Flow (RQ-NSF) \citep{durkan2019neural}. 
After obtaining our likelihood ratio estimates, we can estimate the likelihood of our data by computing $\log p_\textrm{data}(\rvx) \approx \log q(\rvx) = \log r(\rvx) + \log p(\rvx)$. 
We construct our bridge distributions in the latent space of our normalizing flow via the VPSDE interpolation schedule.

We report the likelihoods we obtain via $\dreinf$ in bits per dimension (bpd). 
Additionally, we compare our bpds with both a lower bound estimated via Annealed Importance Sampling (AIS) \citep{neal2001annealed} and a conservative upper bound estimated via the Reverse Annealed Importance Sampling Estimator (RAISE) \citep{burda2015accurate}.
Such comparisons with AIS/RAISE are important because $\log q(\rvx)$ is only an estimate of $\log \pdata$ obtained via $\dreinf$'s approximate normalizing constant. If $\dreinf$ fails to estimate the likelihood ratio $\log r(\rvx)$ accurately, then $\log q(\rvx)$ may be a poor approximation to $\log \pdata$. AIS and RAISE allow us to obtain a more accurate estimate of the intractable normalizing constant by constructing (another) sequence of intermediate distributions between our estimated target distribution $q(\rvx)$ and another proposal distribution $p_1(\rvx)$, which we set to be 
the flow $p(\rvx)$.

As shown in Table~\ref{table:mnist_ebm}, we note that using an infinite number of bridge distributions improves performance on the bpds. 
More importantly, our bpd estimates directly obtained by the output of the score network are very close to those of AIS and RAISE, indicating that our density ratio estimates are accurate even for high dimensional datasets such as MNIST.
This is not necessarily the case for other methods such as TRE.
We refer the reader to Appendix~\ref{app-mnist} for additional details on the experimental setup and likelihood evaluations.

\section{RELATED WORK}
\label{related}
\paragraph{Score-Based Generative Modeling.} 
Our work builds on the growing body of work on score matching \citep{hyvarinen2005estimation,vincent2011connection} and score-based generative models \citep{song2020sliced,song2019generative,song2020improved,song2020score}. Given empirical samples, the goal of score-based generative modeling is to accurately model the data density $p(\rvx)$ by learning its (Stein) \textit{score} $\nabla_\rvx \log p(\rvx)$ \citep{hyvarinen2005estimation,liu2016kernelized}. We notably build upon \citep{song2019generative} to estimate the time scores of the data in addition to the data scores, which allows for accurate DRE.
We additionally establish an interesting connection between time score matching and solving an infinite number of infinitesimal classification problems, which extends the work of \citep{gutmann2012bregman} and \citep{ceylan2018conditional} for (Stein) score matching.

\paragraph{DRE and Importance Sampling.} DRE has its roots in \textit{importance sampling}, which has numerous applications in Bayesian statistics and the approximation of intractable normalizing constants \citep{meng1996simulating,gelman1998simulating,neal2001annealed,fishman2013monte}. In particular, bridge sampling \citep{bennett1976efficient,meng1996simulating} was introduced as a variance reduction technique in MCMC to ``shorten the path" between two densities.
Modern versions of bridge sampling include \citep{rhodes2020telescoping,sinha2020neural}, which also incorporate an element of warping \citep{hoffman2019neutra} via a normalizing flow to further improve performance. \textit{Path sampling} bears the closest resemblance to our method \citep{gelman1998simulating}, in which the discrete bridges of \citep{geyer1994estimating,meng1996simulating} are relaxed to an infinite number as indexed by a continuous value $t \in [0,1]$. However, path sampling estimators are typically not evaluated on a fixed point $\rvx$ as in our use case for DRE.
Another way in which our work differs from traditional parametric importance sampling methods such as AIS is that we do not require explicit parametric forms of the intermediate distributions --- we only require the ability to sample from them. 
Our work most closely mirrors \citep{rhodes2020telescoping}, where we take the number of intermediate distributions to the limit. 
This approach eliminates the need to train multiple classifiers, and makes it easier to incorporate auxiliary information via the data scores to improve DRE in practice.
\section{CONCLUSION}
\label{conclusion}
We introduced $\dreinf$, a novel time score matching framework for DRE. 
We proposed to smoothly interpolate between two densities by specifying an infinite number of bridge distributions, and trained a neural network to estimate the instantaneous rate of change of the log densities (``time scores'') along this path.
After training, we demonstrated that we can leverage black-box numerical integration techniques to efficiently obtain likelihood ratios.
We provide a reference implementation in PyTorch \citep{paszke2019pytorch}, and the codebase for this work is open-sourced at \texttt{https://github.com/ermongroup/dre-infinity}.

However, this work is not without limitations.
Although the method depends on the specification of an interpolation scheme, it is not clear whether there is an optimal way to bridge the two densities together.
Additionally, $\dreinf$ takes longer to converge as $q(\rvx)$ becomes further apart from $p(\rvx)$, though this speaks to the challenging nature of the DRE problem as a whole. 
It would be interesting to investigate whether there is a time-dependent function $\lambda(t)$ such that the time score matching loss corresponds to the maximum likelihood training of a binary classifier \citep{song2021maximum,kingma2021variational}.
Additionally, exploring optimal integration paths between $p(\rvx)$ and $q(\rvx)$ would be exciting future work.

\subsection*{Author Contributions}
Kristy Choi wrote the code, ran the experiments, and wrote the paper. Chenlin Meng helped run the experiments and write the paper. Yang Song designed the project, proposed the theoretical results, and wrote the proofs. Stefano Ermon supervised the project, provided valuable feedback, and helped edit the paper.

\subsection*{Acknowledgements}
We are thankful to Rui Shu and Chris Cundy for providing helpful feedback on the paper.
KC is supported by the NSF GRFP, Stanford Graduate Fellowship, and Two Sigma Diversity PhD Fellowship.
YS is supported by the Apple PhD Fellowship in AI/ML.
This research was supported by NSF (\#1651565, \#1522054, \#1733686), ONR (N00014-19-1-2145), AFOSR (FA9550-19-1-0024), ARO (W911NF2110125), and Amazon AWS.

\bibliographystyle{apalike}
\bibliography{references}

\begin{thebibliography}{}

\bibitem[Abadie and Imbens, 2016]{abadie2016matching}
Abadie, A. and Imbens, G.~W. (2016).
\newblock Matching on the estimated propensity score.
\newblock {\em Econometrica}, 84(2):781--807.

\bibitem[Belghazi et~al., 2018]{belghazi2018mutual}
Belghazi, M.~I., Baratin, A., Rajeshwar, S., Ozair, S., Bengio, Y., Courville,
  A., and Hjelm, D. (2018).
\newblock Mutual information neural estimation.
\newblock In {\em International Conference on Machine Learning}, pages
  531--540. PMLR.

\bibitem[Bennett, 1976]{bennett1976efficient}
Bennett, C.~H. (1976).
\newblock Efficient estimation of free energy differences from monte carlo
  data.
\newblock {\em Journal of Computational Physics}, 22(2):245--268.

\bibitem[Burda et~al., 2015]{burda2015accurate}
Burda, Y., Grosse, R., and Salakhutdinov, R. (2015).
\newblock Accurate and conservative estimates of mrf log-likelihood using
  reverse annealing.
\newblock In {\em Artificial Intelligence and Statistics}, pages 102--110.
  PMLR.

\bibitem[Ceylan and Gutmann, 2018]{ceylan2018conditional}
Ceylan, C. and Gutmann, M.~U. (2018).
\newblock Conditional noise-contrastive estimation of unnormalised models.
\newblock In {\em International Conference on Machine Learning}, pages
  726--734. PMLR.

\bibitem[Chatterjee and Diaconis, 2018]{chatterjee2018sample}
Chatterjee, S. and Diaconis, P. (2018).
\newblock The sample size required in importance sampling.
\newblock {\em The Annals of Applied Probability}, 28(2):1099--1135.

\bibitem[Choi et~al., 2021]{choi2021featurized}
Choi, K., Liao, M., and Ermon, S. (2021).
\newblock Featurized density ratio estimation.
\newblock {\em arXiv preprint arXiv:2107.02212}.

\bibitem[Dormand and Prince, 1980]{dormand1980family}
Dormand, J.~R. and Prince, P.~J. (1980).
\newblock A family of embedded runge-kutta formulae.
\newblock {\em Journal of computational and applied mathematics}, 6(1):19--26.

\bibitem[Durkan et~al., 2019]{durkan2019neural}
Durkan, C., Bekasov, A., Murray, I., and Papamakarios, G. (2019).
\newblock Neural spline flows.
\newblock {\em Advances in Neural Information Processing Systems},
  32:7511--7522.

\bibitem[Fishman, 2013]{fishman2013monte}
Fishman, G. (2013).
\newblock {\em Monte Carlo: concepts, algorithms, and applications}.
\newblock Springer Science \& Business Media.

\bibitem[Gelman and Meng, 1998]{gelman1998simulating}
Gelman, A. and Meng, X.-L. (1998).
\newblock Simulating normalizing constants: From importance sampling to bridge
  sampling to path sampling.
\newblock {\em Statistical science}, pages 163--185.

\bibitem[Geyer, 1994]{geyer1994estimating}
Geyer, C.~J. (1994).
\newblock Estimating normalizing constants and reweighting mixtures.

\bibitem[Goodfellow et~al., 2014]{goodfellow2014generative}
Goodfellow, I.~J., Pouget-Abadie, J., Mirza, M., Xu, B., Warde-Farley, D.,
  Ozair, S., Courville, A., and Bengio, Y. (2014).
\newblock Generative adversarial networks.
\newblock {\em arXiv preprint arXiv:1406.2661}.

\bibitem[Grathwohl et~al., 2018]{grathwohl2018ffjord}
Grathwohl, W., Chen, R.~T., Bettencourt, J., Sutskever, I., and Duvenaud, D.
  (2018).
\newblock Ffjord: Free-form continuous dynamics for scalable reversible
  generative models.
\newblock {\em arXiv preprint arXiv:1810.01367}.

\bibitem[Gretton et~al., 2009]{gretton2009covariate}
Gretton, A., Smola, A., Huang, J., Schmittfull, M., Borgwardt, K., and
  Sch{\"o}lkopf, B. (2009).
\newblock Covariate shift by kernel mean matching.
\newblock {\em Dataset shift in machine learning}, 3(4):5.

\bibitem[Gutmann and Hirayama, 2012]{gutmann2012bregman}
Gutmann, M. and Hirayama, J.-i. (2012).
\newblock Bregman divergence as general framework to estimate unnormalized
  statistical models.
\newblock {\em arXiv preprint arXiv:1202.3727}.

\bibitem[He et~al., 2016]{he2016identity}
He, K., Zhang, X., Ren, S., and Sun, J. (2016).
\newblock Identity mappings in deep residual networks.
\newblock In {\em European conference on computer vision}, pages 630--645.
  Springer.

\bibitem[Ho et~al., 2020]{ho2020denoising}
Ho, J., Jain, A., and Abbeel, P. (2020).
\newblock Denoising diffusion probabilistic models.
\newblock {\em arXiv preprint arXiv:2006.11239}.

\bibitem[Hoffman et~al., 2019]{hoffman2019neutra}
Hoffman, M., Sountsov, P., Dillon, J.~V., Langmore, I., Tran, D., and
  Vasudevan, S. (2019).
\newblock Neutra-lizing bad geometry in hamiltonian monte carlo using neural
  transport.
\newblock {\em arXiv preprint arXiv:1903.03704}.

\bibitem[Hyv{\"a}rinen, 2005]{hyvarinen2005estimation}
Hyv{\"a}rinen, A. (2005).
\newblock Estimation of non-normalized statistical models by score matching.
\newblock {\em Journal of Machine Learning Research}, 6(4).

\bibitem[Johansson et~al., 2018]{johansson2018learning}
Johansson, F.~D., Kallus, N., Shalit, U., and Sontag, D. (2018).
\newblock Learning weighted representations for generalization across designs.
\newblock {\em arXiv preprint arXiv:1802.08598}.

\bibitem[Jordan et~al., 1998]{jordan1998variational}
Jordan, R., Kinderlehrer, D., and Otto, F. (1998).
\newblock The variational formulation of the fokker--planck equation.
\newblock {\em SIAM journal on mathematical analysis}, 29(1):1--17.

\bibitem[Kingma and LeCun, 2010]{kingma2010regularized}
Kingma, D.~P. and LeCun, Y. (2010).
\newblock Regularized estimation of image statistics by score matching.
\newblock In {\em NIPS}, volume 509, page 618.

\bibitem[Kingma et~al., 2021]{kingma2021variational}
Kingma, D.~P., Salimans, T., Poole, B., and Ho, J. (2021).
\newblock Variational diffusion models.
\newblock {\em arXiv preprint arXiv:2107.00630}.

\bibitem[LeCun, 1998]{lecun1998mnist}
LeCun, Y. (1998).
\newblock The mnist database of handwritten digits.
\newblock {\em http://yann. lecun. com/exdb/mnist/}.

\bibitem[Liu et~al., 2016]{liu2016kernelized}
Liu, Q., Lee, J., and Jordan, M. (2016).
\newblock A kernelized stein discrepancy for goodness-of-fit tests.
\newblock In {\em International conference on machine learning}, pages
  276--284. PMLR.

\bibitem[McAllester and Stratos, 2020]{mcallester2020formal}
McAllester, D. and Stratos, K. (2020).
\newblock Formal limitations on the measurement of mutual information.
\newblock In {\em International Conference on Artificial Intelligence and
  Statistics}, pages 875--884. PMLR.

\bibitem[Meng and Wong, 1996]{meng1996simulating}
Meng, X.-L. and Wong, W.~H. (1996).
\newblock Simulating ratios of normalizing constants via a simple identity: a
  theoretical exploration.
\newblock {\em Statistica Sinica}, pages 831--860.

\bibitem[Menon and Ong, 2016]{menon2016linking}
Menon, A. and Ong, C.~S. (2016).
\newblock Linking losses for density ratio and class-probability estimation.
\newblock In {\em International Conference on Machine Learning}, pages
  304--313. PMLR.

\bibitem[Neal, 1993]{neal1993probabilistic}
Neal, R.~M. (1993).
\newblock {\em Probabilistic inference using Markov chain Monte Carlo methods}.
\newblock Department of Computer Science, University of Toronto Toronto, ON,
  Canada.

\bibitem[Neal, 2001]{neal2001annealed}
Neal, R.~M. (2001).
\newblock Annealed importance sampling.
\newblock {\em Statistics and computing}, 11(2):125--139.

\bibitem[Nguyen et~al., 2007]{nguyen2007estimating}
Nguyen, X., Wainwright, M.~J., and Jordan, M.~I. (2007).
\newblock Estimating divergence functionals and the likelihood ratio by
  penalized convex risk minimization.
\newblock In {\em NIPS}, pages 1089--1096.

\bibitem[Nichol and Dhariwal, 2021]{nichol2021improved}
Nichol, A. and Dhariwal, P. (2021).
\newblock Improved denoising diffusion probabilistic models.
\newblock {\em arXiv preprint arXiv:2102.09672}.

\bibitem[Nowozin et~al., 2016]{nowozin2016f}
Nowozin, S., Cseke, B., and Tomioka, R. (2016).
\newblock f-gan: Training generative neural samplers using variational
  divergence minimization.
\newblock {\em arXiv preprint arXiv:1606.00709}.

\bibitem[{\O}ksendal, 2003]{oksendal2003stochastic}
{\O}ksendal, B. (2003).
\newblock Stochastic differential equations.
\newblock In {\em Stochastic differential equations}, pages 65--84. Springer.

\bibitem[Oord et~al., 2018]{oord2018representation}
Oord, A. v.~d., Li, Y., and Vinyals, O. (2018).
\newblock Representation learning with contrastive predictive coding.
\newblock {\em arXiv preprint arXiv:1807.03748}.

\bibitem[Owen, 2013]{owen2013monte}
Owen, A.~B. (2013).
\newblock Monte carlo theory, methods and examples.

\bibitem[Paszke et~al., 2019]{paszke2019pytorch}
Paszke, A., Gross, S., Massa, F., Lerer, A., Bradbury, J., Chanan, G., Killeen,
  T., Lin, Z., Gimelshein, N., Antiga, L., et~al. (2019).
\newblock Pytorch: An imperative style, high-performance deep learning library.
\newblock {\em Advances in neural information processing systems}, 32.

\bibitem[Poole et~al., 2019]{poole2019variational}
Poole, B., Ozair, S., Van Den~Oord, A., Alemi, A., and Tucker, G. (2019).
\newblock On variational bounds of mutual information.
\newblock In {\em International Conference on Machine Learning}, pages
  5171--5180. PMLR.

\bibitem[Ramachandran et~al., 2017]{ramachandran2017searching}
Ramachandran, P., Zoph, B., and Le, Q.~V. (2017).
\newblock Searching for activation functions.
\newblock {\em arXiv preprint arXiv:1710.05941}.

\bibitem[Rhodes et~al., 2020]{rhodes2020telescoping}
Rhodes, B., Xu, K., and Gutmann, M.~U. (2020).
\newblock Telescoping density-ratio estimation.
\newblock {\em Advances in Neural Information Processing Systems}, 33.

\bibitem[Ronneberger et~al., 2015]{ronneberger2015u}
Ronneberger, O., Fischer, P., and Brox, T. (2015).
\newblock U-net: Convolutional networks for biomedical image segmentation.
\newblock In {\em International Conference on Medical image computing and
  computer-assisted intervention}, pages 234--241. Springer.

\bibitem[Salimans and Ho, 2021]{salimans2021should}
Salimans, T. and Ho, J. (2021).
\newblock Should ebms model the energy or the score?
\newblock In {\em Energy Based Models Workshop-ICLR 2021}.

\bibitem[Shalit et~al., 2017]{shalit2017estimating}
Shalit, U., Johansson, F.~D., and Sontag, D. (2017).
\newblock Estimating individual treatment effect: generalization bounds and
  algorithms.
\newblock In {\em International Conference on Machine Learning}, pages
  3076--3085. PMLR.

\bibitem[Sinha et~al., 2020]{sinha2020neural}
Sinha, A., O'Kelly, M., Tedrake, R., and Duchi, J.~C. (2020).
\newblock Neural bridge sampling for evaluating safety-critical autonomous
  systems.
\newblock {\em Advances in Neural Information Processing Systems}, 33.

\bibitem[Sohl-Dickstein et~al., 2015]{sohl2015deep}
Sohl-Dickstein, J., Weiss, E., Maheswaranathan, N., and Ganguli, S. (2015).
\newblock Deep unsupervised learning using nonequilibrium thermodynamics.
\newblock In {\em International Conference on Machine Learning}, pages
  2256--2265. PMLR.

\bibitem[Song and Ermon, 2019a]{song2019understanding}
Song, J. and Ermon, S. (2019a).
\newblock Understanding the limitations of variational mutual information
  estimators.
\newblock {\em arXiv preprint arXiv:1910.06222}.

\bibitem[Song et~al., 2021a]{song2021maximum}
Song, Y., Durkan, C., Murray, I., and Ermon, S. (2021a).
\newblock Maximum likelihood training of score-based diffusion models.
\newblock {\em arXiv e-prints}, pages arXiv--2101.

\bibitem[Song and Ermon, 2019b]{song2019generative}
Song, Y. and Ermon, S. (2019b).
\newblock Generative modeling by estimating gradients of the data distribution.
\newblock {\em arXiv preprint arXiv:1907.05600}.

\bibitem[Song and Ermon, 2020]{song2020improved}
Song, Y. and Ermon, S. (2020).
\newblock Improved techniques for training score-based generative models.
\newblock {\em arXiv preprint arXiv:2006.09011}.

\bibitem[Song et~al., 2020]{song2020sliced}
Song, Y., Garg, S., Shi, J., and Ermon, S. (2020).
\newblock Sliced score matching: A scalable approach to density and score
  estimation.
\newblock In {\em Uncertainty in Artificial Intelligence}, pages 574--584.
  PMLR.

\bibitem[Song et~al., 2021b]{song2020score}
Song, Y., Sohl-Dickstein, J., Kingma, D.~P., Kumar, A., Ermon, S., and Poole,
  B. (2021b).
\newblock Score-based generative modeling through stochastic differential
  equations.
\newblock In {\em International Conference on Learning Representations}.

\bibitem[Sugiyama et~al., 2008]{sugiyama2008direct}
Sugiyama, M., Suzuki, T., Nakajima, S., Kashima, H., von B{\"u}nau, P., and
  Kawanabe, M. (2008).
\newblock Direct importance estimation for covariate shift adaptation.
\newblock {\em Annals of the Institute of Statistical Mathematics},
  60(4):699--746.

\bibitem[Tancik et~al., 2020]{tancik2020fourier}
Tancik, M., Srinivasan, P.~P., Mildenhall, B., Fridovich-Keil, S., Raghavan,
  N., Singhal, U., Ramamoorthi, R., Barron, J.~T., and Ng, R. (2020).
\newblock Fourier features let networks learn high frequency functions in low
  dimensional domains.
\newblock {\em arXiv preprint arXiv:2006.10739}.

\bibitem[Vincent, 2011]{vincent2011connection}
Vincent, P. (2011).
\newblock A connection between score matching and denoising autoencoders.
\newblock {\em Neural computation}, 23(7):1661--1674.

\bibitem[Wu and He, 2018]{wu2018group}
Wu, Y. and He, K. (2018).
\newblock Group normalization.
\newblock In {\em Proceedings of the European conference on computer vision
  (ECCV)}, pages 3--19.

\bibitem[Yamada et~al., 2013]{yamada2013relative}
Yamada, M., Suzuki, T., Kanamori, T., Hachiya, H., and Sugiyama, M. (2013).
\newblock Relative density-ratio estimation for robust distribution comparison.
\newblock {\em Neural computation}, 25(5):1324--1370.

\bibitem[Yao et~al., 2020]{yao2020adaptive}
Yao, Y., Cademartori, C., Vehtari, A., and Gelman, A. (2020).
\newblock Adaptive path sampling in metastable posterior distributions.
\newblock {\em arXiv preprint arXiv:2009.00471}.

\end{thebibliography}
\raggedbottom
\pagebreak



\clearpage
\appendix

\thispagestyle{empty}

\onecolumn \makesupplementtitle
\renewcommand{\thesubsection}{\Alph{subsection}}

\newtheorem*{P1}{Proposition~\ref{sum_to_integral}}
\newtheorem*{P2}{Proposition~\ref{prop:score_clf}}
\newtheorem*{P3}{Proposition~\ref{prop:opt_clf}}
\newtheorem*{P4}{Proposition~\ref{prop:time_objective}}
\newtheorem*{T1}{Theorem~\ref{thm:objective}}

\label{appendix}

\subsection{Detailed Derivations of Theoretical Results}
In this section, we provide a more careful treatment of the relevant derivations in the main text.

\subsubsection{Bridge Sampling to Path Sampling}
\label{path_sampling}
The identity for converting bridge sampling to path sampling in \cref{sum_to_integral} is well known
\citep{gelman1998simulating,owen2013monte,yao2020adaptive}, and we include it here for completeness.
\begin{P1}
Let $\log r(\rvx)$ denote the log density ratio between the two densities $p_0(\rvx)$ and $p_1(\rvx)$.
When $T \rightarrow \infty$, we have the following:
\begin{equation}
    \begin{split}
    \log r(\rvx) = \log \frac{p_0(\rvx)}{p_1(\rvx)} &= \sum_{t=1}^{T} \log \left( \frac{p_{(t-1)/T}(\rvx)}{p_{t/T}(\rvx)} \right)
    = \int_{1}^0 \frac{\partial}{\partial \lambda} \log p_\lambda(\rvx) d\lambda 
    \end{split}
\end{equation}
\end{P1}
\begin{proof}
\begin{align*}
    \log \frac{p_0(\rvx)}{p_1(\rvx)} &= \log p_0(\rvx) - \log p_1(\rvx) \\
    &= \left( \log p_0(\rvx) - \log p_{1/T}(\rvx) \right) + \left( \log p_{1/T}(\rvx) - \log p_{2/T}(\rvx) \right) + \cdots + \left( \log p_{(T-1)/T}(\rvx) - \log p_1(\rvx) \right) \\
    &= \sum_{t=1}^T \log \left( \frac{p_{(t-1)/T}(\rvx)}{p_{t/T}(\rvx)} \right) \\
    &= \sum_{t=1}^T \log \left(1 + \frac{p_{(t-1)/T}(\rvx) - p_{t/T}(\rvx)}{p_{t/T}(\rvx)} \right)\\
    & \approx \sum_{t=1}^T \left( \frac{p_{(t-1)/T}(\rvx) - p_{t/T}(\rvx)}{p_{t/T}(\rvx)} \right) \\
    &= \lim_{T \rightarrow \infty}  \sum_{t=1}^T \frac{d}{d \lambda} \log p_\lambda(\rvx) \vert_{\lambda=t/T} \\
    &= \int_{1}^0 \log p_\lambda(\rvx) d\lambda
\end{align*}
\end{proof}

\subsubsection{Form of Optimal Infinitesimal Classifier}
For completeness, we restate Proposition~\ref{prop:score_clf} prior to providing the proof.
\begin{P2}
When $T \rightarrow \infty$, the form of the Bayes-optimal classifier between two adjacent bridge distributions $p_{t/T}(\rvx)$ and $p_{(t+1)/T}(\rvx)$ for any $t \in [0,1]$ becomes:
\begin{equation}
    \vh_{\vtheta^*}(\rvx, t) = \frac{1}{2} + \frac{1}{4} \left(\dt \log p_t(\rvx)\right)\Delta t + o(\Delta t).
\end{equation}
\end{P2}
\begin{proof}
Recall that it is trained by optimizing the following cross-entropy loss:
\begin{align*}
    \vtheta^* = \argmax_\vtheta \mathbb{E}_{p_{(t-1)/T}(\rvx)}[\log (1 - \vh_\vtheta(\rvx, t/T))] + \mathbb{E}_{p_{t/T}(\rvx)}[\log \vh_\vtheta(\rvx, t/T)],
\end{align*}
where $\vh_\vtheta(\rvx, t/T) \in [0,1]$ is a binary classifier. By calculus of variations, we can derive that for the optimal model parameter $\vtheta^*$,
\begin{align*}
    \vh_{\vtheta^*}(\rvx, t/T) = \sigma(\log p_{t/T}(\rvx) - \log p_{(t-1)/T}(\rvx)),
\end{align*}
where $\sigma(\rvx) = \frac{1}{1+e^{-\rvx}}$ is the sigmoid function. 
Thus when $T\to\infty$, we clearly have that:
\begin{align*}
    \vh_{\vtheta^*}(\rvx, t) = \frac{1}{2} + \frac{1}{4} \left(\dt \log p_t(\rvx)\right)\Delta t + o(\Delta t).
\end{align*}
where $\Delta t = \frac{1}{T}$.
\end{proof}

\subsubsection{Derivation of the Time Score from Infinitesimal Binary Classification}
\label{pf-opt-clf}
For completeness, we restate Proposition~\ref{prop:opt_clf} prior to providing the proof.
\begin{P3}
Let $\Delta t = 1/T$ and parameterize the binary classifier as $\vh_\vtheta(\rvx, t) = \frac{1}{2} + \frac{1}{4} \vs_\vtheta^{\textrm{time}}(\rvx, t)\Delta t$. Then from the binary cross-entropy objective, we can derive:
\begin{equation}
\begin{split}
    &\arg \max_\theta \mathbb{E}_{p_{t}(\rvx)}[\log (1 - \vh_\vtheta(\rvx, t))] + \mathbb{E}_{p_{t+\Delta t}(\rvx)}[\log \vh_\vtheta(\rvx, t)]
    = \arg \max_\theta -\frac{1}{4} (\Delta t)^2 \mbb{E}_{p_t(\rvx)}\left[\norm{\vs_\vtheta^{\textrm{time}}(\rvx, t) - \dt \log p_t(\rvx)}_2^2\right] + o((\Delta t)^2)
\end{split}
\end{equation}
\end{P3}
\begin{proof}
From the definition of the binary cross entropy loss, we have:
\begin{align*}
    \arg \max_\theta \mathbb{E}_{p_{t}(\rvx)}[\log (1 - \vh_\vtheta(\rvx, t))] + \mathbb{E}_{p_{t+\Delta t}(\rvx)}[\log \vh_\vtheta(\rvx, t)]
    &= \arg \max_\theta -2\log 2 + \frac{1}{2}\Delta t \int (p_{t+\Delta t}(\rvx) - p_t(\rvx))\vs_\vtheta^{\textrm{time}}(\rvx, t) \ud \rvx \\
    & -\frac{1}{4} (\Delta t)^2 \mathbb{E}_{p_t(\rvx)}[\vs_\vtheta^{\textrm{time}}(\rvx, t)^2]+ o((\Delta t)^2)\\
    =& \arg \max_\theta -2\log 2 + \frac{1}{2}(\Delta t)^2 \mbb{E}_{p_t(\rvx)}\left[\dt \log p_t(\rvx) \vs_\vtheta^{\textrm{time}}(\rvx, t)\right] \\
    & -\frac{1}{4} (\Delta t)^2 \mathbb{E}_{p_t(\rvx)}[\vs_\vtheta^{\textrm{time}}(\rvx, t)^2]+ o((\Delta t)^2)\\
    =& \arg \max_\theta -\frac{1}{4} (\Delta t)^2 \mbb{E}_{p_t(\rvx)}\left[\norm{\vs_\vtheta^{\textrm{time}}(\rvx, t) - \dt \log p_t(\rvx)}_2^2\right] \\ &+ o((\Delta t)^2)
\end{align*}
\end{proof}

\subsubsection{Time score matching objective}
\label{pf-time-prop}
We provide a more detailed derivation of the time-wise score matching objective in \cref{time_obj_first} below.
\begin{P4}
Under certain regularity conditions, the optimal solution $\theta^{*}$ of \cref{time_obj_first} is the same as the optimal solution of:
\begin{equation}
\begin{split}
        &\mathcal{L_{\text{time}}(\theta)} = 2 \E_{q(\rvx)}[\lambda(0) s_\vtheta^{\textrm{time}}(\rvx, 0)] - 2\E_{p(\rvx)}[\lambda(1) s_\vtheta^{\textrm{time}}(\rvx, 1)] 
    +\E_{p(t)}\E_{p_t(\rvx)}\bigg[2 \lambda(t) \dt s_\vtheta^{\textrm{time}}(\rvx, t) + 2\lambda'(t) s_\vtheta^{\textrm{time}}(\rvx, t) + \lambda(t) s_\vtheta^{\textrm{time}}(\rvx, t)^2\bigg]
\end{split}
\end{equation}
\end{P4}
\begin{proof}
To see this, we expand out the square and use the Leibniz integral rule.
\begin{align*}
    &\E_{p(t)}\E_{p_t(\rvx)}\bigg[\lambda(t) \left( \dt \log p_t(\rvx) - s_\vtheta^{\textrm{time}}(\rvx, t) \right)^2 \bigg] \\
    =&\E_{p(t)}\E_{p_t(\rvx)}\bigg[\lambda(t) \bigg(\dt \log p_t(\rvx)\bigg)^2 - 2\lambda(t) s_\vtheta^{\textrm{time}}(\rvx, t)\dt \log p_t(\rvx) + \lambda(t) s_\vtheta^{\textrm{time}}(\rvx, t)^2\bigg]\\
    =& \E_{p(t)}\E_{p_t(\rvx)}\bigg[ - 2 \lambda(t) s_\vtheta^{\textrm{time}}(\rvx, t)\dt \log p_t(\rvx) + \lambda(t) s_\vtheta^{\textrm{time}}(\rvx, t)^2\bigg] + \textnormal{const.}\\
    =& \E_{p(t)}\E_{p_t(\rvx)}\bigg[ - 2 \lambda(t) s_\vtheta^{\textrm{time}}(\rvx, t)\dt \log p_t(\rvx) \bigg] +\E_{p(t)}\E_{p_t(\rvx)}[\lambda(t) s_\vtheta^{\textrm{time}}(\rvx, t)^2] + \textnormal{const.}\\
    =& \int_0^1\int  - 2 \lambda(t) s_\vtheta^{\textrm{time}}(\rvx, t)\dt \log p_t(\rvx)p_t(\rvx) \ud \rvx \ud t +\E_{p(t)}\E_{p_t(\rvx)}[\lambda(t) s_\vtheta^{\textrm{time}}(\rvx, t)^2] + \textnormal{const.}\\
    =& -2 \int_0^1\int  \lambda(t) s_\vtheta^{\textrm{time}}(\rvx, t)\frac{\partial p_t(\rvx)}{\partial t} \ud \rvx \ud t +\E_{p(t)}\E_{p_t(\rvx)}[\lambda(t) s_\vtheta^{\textrm{time}}(\rvx, t)^2] + \textnormal{const.}\\
    =& 2\int [\lambda(0) p_0(\rvx) s_\vtheta^{\textrm{time}}(\rvx, 0) - \lambda(t) 2p_1(\rvx) s_\vtheta^{\textrm{time}}(\rvx, 1)] \ud \rvx \\
    &+ 2\int_0^1\int  p_t(\rvx) \bigg[ \lambda(t) \dt s_\vtheta^{\textrm{time}}(\rvx, t) + \lambda'(t) s_\vtheta^{\textrm{time}}(\rvx, t)\bigg] \ud \rvx \ud t +\E_{p(t)}\E_{p_t(\rvx)}[\lambda(t) s_\vtheta^{\textrm{time}}(\rvx, t)^2] + \textnormal{const.}\\
    =& 2 \E_{q(\rvx)}[\lambda(0) s_\vtheta^{\textrm{time}}(\rvx, 0)] - 2\E_{p(\rvx)}[\lambda(1) s_\vtheta^{\textrm{time}}(\rvx, 1)]\\
    & +\E_{p(t)}\E_{p_t(\rvx)}\bigg[2 \lambda(t) 
    \dt s_\vtheta^{\textrm{time}}(\rvx, t) + 2\lambda'(t) s_\vtheta^{\textrm{time}}(\rvx, t) + \lambda(t) s_\vtheta^{\textrm{time}}(\rvx, t)^2\bigg] + \textnormal{const.}
\end{align*}
\end{proof}
The optimal time score model, denoted by $s_{\vtheta^\ast}^{\textrm{time}}(\rvx, t)$, satisfies $s_{\vtheta^*}^{\textrm{time}}(\rvx, t) \approx \dt \log p_t(\rvx)$. Therefore, the log-density-ratio can be estimated by
\begin{align*}
    \log r(\rvx) \approx \int_{1}^0 s_{\vtheta^*}^{\textrm{time}}(\rvx, t) \ud t.
\end{align*}

\subsubsection{Joint Score Matching Objective}
The assumptions needed for this proof are largely adapted from \citep{song2020sliced}.
\label{app-joint-proof}
\begin{T1}
Assume that the vector-valued score function learned by the joint score network $\vs_\vtheta^{\textrm{joint}}(\rvx, t)$ and the true data scores $\nabla_\rvx \log p_t(\rvx)$ are differentiable, and satisfy $\bbE[\Vert \vs_\vtheta^{\textrm{joint}}(\rvx, t) \Vert_2^2] < \infty$ and $\bbE[\Vert \nabla_\rvx \log p_t(\rvx) \Vert_2^2] < \infty$. 
Additionally, we assume: (1) identifiability---the model family $\{p_m(\rvx; \theta) | \theta \in \Theta\}$ is well-specified; and 
(2) that the score model satisfies some boundary conditions, e.g. $\forall \theta \in \Theta, \lim_{\Vert \rvx \Vert \rightarrow \infty} s_\vtheta^{\textrm{joint}}(\rvx, t) \pdata = 0$.
We also assume that the projection vectors $\rvv \sim p(\rvv) = \mcal{N}(\bm{0}, \mI)$. 
Then, the solution to the optimization problem in \cref{eqn:joint_obj} can be written as follows:
\begin{equation}
\begin{split}
    \theta^{*} &= \argmin_{\theta} \textrm{}
    \E_{p(t)}\E_{p_{t}(\rvx)}\E_{p({\rvv})}\bigg[
    \frac{1}{2}\lambda(t) \norm{\vs_\vtheta^{\textrm{joint}}(\rvx, t)[\rvx]}_2^2 + \lambda(t) \rvv\tran \nabla_{\rvx} \vs_\vtheta^{\textrm{joint}}(\rvx, t)[\rvx] \rvv \\
    &+ \lambda(t) \frac{\partial}{\partial t}\vs_\vtheta^{\textrm{joint}}(\rvx, t)[t] + \lambda'(t) \vs_\vtheta^{\textrm{joint}}(\rvx, t)[t]\bigg] \\
    &+ \E_{p_0(\rvx)}[\lambda(0) \vs_\vtheta^{\textrm{joint}}(\rvx, 0)[t]] - \E_{p_1(\rvx)}[\lambda(1) \vs_\vtheta^{\textrm{joint}}(\rvx, 1)[t]].
\end{split}
\end{equation}
\end{T1}


\begin{proof}
The proof involves expanding out the square and using integration by parts as in \citep{hyvarinen2005estimation}.
\begin{equation}
\begin{split}
    \mathcal{L_{\text{joint}}(\theta)} =& \E_{p(t)}\E_{p_{t}(\rvx)}\left[ \frac{1}{2}\lambda(t)\norm{\nabla_{[\rvx; t]}\log p(\rvx, t) - \vs_\vtheta^{\textrm{joint}}(\rvx, t)}_2^2 \right]\\
    =&\E_{p(t)}\E_{p_{t}(\rvx)}\left[ \frac{1}{2} \lambda(t) \norm{\vs_\vtheta^{\textrm{joint}}(\rvx, t)[\rvx]}_2^2 + \lambda(t) \operatorname{tr}(\mJ_{\vs_\vtheta}(\rvx, t)) + \lambda'(t) \vs_\vtheta^{\textrm{joint}}(\rvx, t)[t] \right] \\
    &+ \E_{p_0(\rvx)}[\lambda(0) \vs_\vtheta^{\textrm{joint}}(\rvx, 0)[t]] - \E_{p_1(\rvx)}[\lambda(1) \vs_\vtheta^{\textrm{joint}}(\rvx, 1)[t]] + \text{const.}\\
    =&\E_{p(t)}\E_{p_{t}(\rvx)}\E_{p(\rvv)}\left[ \frac{1}{2}\lambda(t) \norm{\vs_\vtheta^{\textrm{joint}}(\rvx, t)[\rvx]}_2^2 + \lambda(t) \rvv\tran \nabla_{\rvx} \vs_\vtheta^{\textrm{joint}}(\rvx, t)[\rvx] \rvv + \lambda(t) \frac{\partial}{\partial t}\vs_\vtheta^{\textrm{joint}}(\rvx, t)[t] + \lambda'(t) \vs_\vtheta^{\textrm{joint}}(\rvx, t)[t]\right] \\
    &+ \E_{p_0(\rvx)}[\lambda(0) \vs_\vtheta^{\textrm{joint}}(\rvx, 0)[t]] - \E_{p_1(\rvx)}[\lambda(1) \vs_\vtheta^{\textrm{joint}}(\rvx, 1)[t]] + \text{const}.
\end{split}
\end{equation}
\end{proof} 
We note that for the joint training objective, the optimal score model satisfies $\vs_{\vtheta^*}^{\textrm{joint}}(\rvx, t) \approx [\nabla_\rvx \log p_{t}(\rvx); \dt \log p_{t}(\rvx)]$. This is because $\vs_{\vtheta^*}^{\textrm{joint}}(\rvx, t) \approx \nabla_{[\rvx;t]} \log p(\rvx, t)$ and
\begin{align*}
    \nabla_\rvx \log p(\rvx, t) &= \nabla_\rvx \log p_{t}(\rvx) + \nabla_\rvx \log p(t) = \nabla_\rvx \log p_{t}(\rvx)\\
    \dt \log p(\rvx, t) &= \dt \log p_{t}(\rvx) + \dt \log p(t) = \dt \log p_{t}(\rvx),
\end{align*}
since $p(t)$ does not depend on $\rvx$ and is a uniform distribution.

\subsection{Pseudocode for Training and Inference}
\label{app-pseudo}
We provide pseudocode for training the time score model using \cref{time_obj_first}.
\begin{algorithm}[!ht]
  \caption{Time Score Matching}
  \label{alg:dreinf}
  \textbf{Input:} Datasets $\{\Dp, \Dq\}$, time score model $\vs_\vtheta^{\textrm{time}}(\rvx, t)$, interpolation procedure \texttt{interpolate}, weighting function $\lambda: [0,1] \rightarrow \mathbb{R}_+$\\
\begin{algorithmic}[1]
\STATE Sample $t \sim \mcal{U}[0,1]$\\
\STATE Sample $\rvx \sim \Dq$, $\rvy \sim \Dp$\\
\STATE Interpolate $\rvx_t \leftarrow \texttt{interpolate}(\rvx, \rvy, t)$ \\
\STATE $\hat{\mathcal{L}}(\theta) \leftarrow \lambda(t) \dt s_\vtheta^{\textrm{time}}(\rvx_t, t) + 2\lambda'(t) s_\vtheta^{\textrm{time}}(\rvx_t, t) + \lambda(t) s_\vtheta^{\textrm{time}}(\rvx_t, t) + 2\lambda(0) s_\vtheta^{\textrm{time}}(\rvx, 0) - 2\lambda(1) s_\vtheta^{\textrm{time}}(\rvy, 1)$\\
 \STATE {\bfseries return} $\hat{\mathcal{L}}(\theta)$ \\
\end{algorithmic}
\end{algorithm}

Next, we provide pseudocode for computing the density ratios via any black-box numerical integration method.
\begin{algorithm}[!ht]
  \caption{Density Ratio Estimation}
  \label{alg:dreinf}
  \textbf{Input:} Time score model $\vs_\vtheta^{\textrm{time}}$, minibatch of samples $\rvx$, start time $t_0=1$, end time $t_1=0$, initial condition $\rvy_0=\bm{0}$\\
\begin{algorithmic}[1]
\STATE $s_\vtheta^{\rvx} \leftarrow s_\vtheta^{\textrm{time}}(\rvx, \cdot)$ \\
\STATE $\log r(\rvx) \leftarrow$ \texttt{integrate($s_\vtheta^{\rvx}$, $(t_0,t_1)$, $\rvy_0$)} \\
 \STATE {\bfseries return} $\log r(\rvx)$ \\
\end{algorithmic}
\end{algorithm}

\subsection{Structured Interpolations via Stochastic Differential Equations (SDEs)}
We briefly mention a special case of $\dreinf$'s interpolation mechanism where the analytical form of $p_t(\rvx)$ is tractable.
One such example is a diffusion process, where the data generating process of $q(\rvx)$ is represented as a Markov chain that transforms a simple distribution $p_T(\rvx) \equiv p(\rvx)$
into a target distribution $p_0(\rvx) \equiv q(\rvx)$ \citep{sohl2015deep,ho2020denoising,song2020score}. 
This sequential procedure can be described as the solution to an It$\hat{\textrm{o}}$ stochastic differential equation (SDE):
\[
\dd\rvx = \bff(\rvx, t)\dd t + g(t)\dd \bw
\]
where $\bw$ represents Brownian motion, $\bff(\cdot,t): \mathbb{R}^D \rightarrow \mathbb{R}^D$ is the drift coefficient of $\rvx(t)$, and $g(\cdot): \mathbb{R} \rightarrow \mathbb{R}$ is the diffusion coefficient of $\rvx(t)$, where $\rvx(t) \sim p_t(\rvx)$. 

For learning settings where our data follows a known SDE, 
we can leverage the Fokker-Planck equation \citep{jordan1998variational,oksendal2003stochastic} to transform the data scores $\nabla_\rvx \log p_t(\rvx)$ into time scores \emph{without training an additional model}.
The Fokker-Planck equation describes the time-evolution of the probability density $p_t(\rvx)$ associated with the SDE:
\begin{equation}
\label{fokker-planck}
\begin{split}
    \frac{\partial}{\partial t}\log p_t(\rvx) &= -\nabla \cdot \bff - \bff \tran \nabla_\rvx \log p_t(\rvx) \\
    &+ g^2(t) \left[ ||\nabla_\rvx \log p_t(\rvx) ||^2_2 + \textrm{tr}(\nabla_\rvx^2 \log p_t(\rvx)) \right]
\end{split}
\end{equation}
where $\bff$ and $g$ are well-defined for tractable SDEs. 
We can train a data score network $s_\vtheta^{\textrm{data}}(\rvx, t)$ to approximate $\nabla_\rvx \log p_t(\rvx)$ via sliced score matching (SSM) \citep{song2020sliced} or denoising score matching (DSM) \citep{vincent2011connection,song2019generative}. 


\subsection{Architecture and Implementation Details}
\label{app-arch}
We provide further details on the importance weighting scheme and the time score architecture discussed in Section~\ref{impl-design}.

\subsubsection{Variance Reduction via Polynomial Interpolation with a Loss History Buffer}
\label{app:loss_history}
In our empirical evaluations, we experimented with various approaches for learning the proper weighting function $p_{\text{iw}}(t) \approx \lambda(t): [0,1] \rightarrow \mathbb{R}_+$ to reduce the variance in our training objectives. We found that estimating the weights by maintaining a history of the $B>0$ most recent loss values \citep{nichol2021improved} led to the most significant performance improvements. We used this loss history approach for the energy-based modeling experiments in MNIST, it was not necessary for our synthetic experiments.

We experimented with buffer sizes of $B=\{10, 100\}$, and batch sizes of \{64, 128, 256, 500\}. For each minibatch $t \sim \mathcal{U}[0,1]$ sampled during training, we sorted the timescales in ascending order before computing the loss and storing the corresponding values into the buffer. This made the batch size very important, as it served as a discrete approximation to the way in the history was used to compute the weights in \citep{nichol2021improved}. After obtaining an initial estimate of the weights as in \citep{nichol2021improved}, we fit a ridge regression model using the stored time and weight values with the default settings for \texttt{PolynomialFeatures} in \texttt{scikit-learn} (polynomial degree 4 and a regularization coefficient of $0.001$). This regression model was then used as an interpolation mechanism for returning the corresponding weight values for new values of $t$ seen during training.

To avoid settings where the loss weighting would return negative values for $\alpha(t)$, we returned the absolute values of the interpolated weights before applying them in our loss function. For our MNIST experiments, we found that a buffer size of $B=100$ and a batch size of $500$ worked the best for the Gaussian noise and Gaussian copula noise models. For the RQ-NSF noise model, this interpolation mechanism did not improve performance (and thus we used the original VPSDE weighting scheme instead).

\subsubsection{Time Score Network Architecture} 
\label{time_arch}
When designing the time score network architecture for more complex datasets, we found that both sinusoidal positional embeddings \citep{ho2020denoising} and Fourier embeddings \citep{tancik2020fourier,song2020score,kingma2021variational} commonly used in the literature led to training instabilities when computing $\dt \vs_{\vtheta}^{\textrm{time}}(\rvx, t)$. 
We hypothesize that this is due to the periodic nature of the sine and cosine functions, causing gradient information with respect to $t$ to oscillate during training.
Therefore, we fed the time-conditioning signal into a single hidden-layer Multilayer Perception (MLP) with Tanh activation functions prior to combining it with the input features.
We composed this time embedding module with a convolutional U-Net architecture \citep{ronneberger2015u}, which gave us the biggest performance boost, for our experiments.
We defer additional details to Appendix~\ref{app:loss_history}.

In both the time-wise and joint score matching objectives in \cref{time_obj_first} and \cref{eqn:joint_obj}, we must backpropagate through the network with respect to the time-conditioning signal $t$. This requires care in designing the embedding mechanism as well as the network architecture to avoid training instabilities. Below, we list additional details as well as some empirical observations that led to good performance in practice:
\begin{enumerate}
    \item For the time embeddings, we used a single-hidden layer MLP of the following form: \texttt{Linear(1, 256) $\rightarrow$ \texttt{Tanh} $\rightarrow$ Linear(256, 256) $\rightarrow$ \texttt{Tanh} $\rightarrow$ Linear(256, 256)}.
    \item The \texttt{Swish} activation function \citep{ramachandran2017searching} and Group Normalization \citep{wu2018group} led to the most stable training in our score networks for the MNIST experiments. For all other synthetic experiments, we used the \texttt{ELU} activation function. In general, we found that commonly used activation functions such as \texttt{ReLU} and \texttt{LeakyReLU} hurt performance, as backpropagating through the network during training would zero out gradients. 
    \item We found that architecture backbones based on ResNets \citep{he2016identity} led to unstable training. 
    \item For our model architecture, we used a standard convolutional U-Net with channels of increasing resolution [64, 128, 256, 512]. The details of this architecture can be found in Table~\ref{table:unet_arch}. We note that after each convolution, the Dense activation block is applied to the time embedding and added to the convolved input feature. Then, the output of this operation is passed through a Group Normalization layer and the \texttt{Swish} activation function.
    \item As in a standard U-Net: the output of the 3rd convolutional block (combined with the time embedding, plus normalization/activation) is concatenated with the input to \texttt{tconv3}, the output of the 2nd convolutional block is concatenated with the previous output into into \texttt{tconv2}, etc.
    \item In our convolutional U-Net score network, we incorporated the outputs of the time-conditioning MLP module via \texttt{Dense activation} blocks. This block is also a single-hidden layer MLP with with \texttt{Tanh} activations of the following structure: \texttt{Linear(256, 32) $\rightarrow$ \texttt{Tanh} $\rightarrow$ Linear(32, 32) $\rightarrow$ \texttt{Tanh} $\rightarrow$ Linear(32, U-Net channel)}, where 256 corresponds to the output size of the time-embedding module.
\end{enumerate}

\subsection{Leveraging the Numerical Integrator for Density Ratio Estimation}
After training a time-conditioned score network with \cref{time_obj_first}, it is straightforward to see that $\log r(\rvx)$ can be obtained via the following formula:
\begin{align*}
    \log r(\rvx) = \int_{1}^0 \dt \log p_t(\rvx)\ud t \approx \int_{1}^0 \vs_{\vtheta}^{\textrm{time}}(\rvx, t) \ud t. 
\end{align*}
The integration over all intermediate time scores in \cref{eqn:dr1} can be computed
using any existing numerical integrator. 
In our experiments, we leverage a black-box ODE solver to perform the integration, though we emphasize that using an ODE solver is not strictly necessary.
The ODE solver determines the timesteps $t$ we should query along the trajectory as we obtain our density ratio estimates, which eliminates the need to hand-tune $T$ as in \cref{eq:tre}.
For computing the likelihood ratios as in \cref{eqn:dr1} in all our experiments, we follow \citep{grathwohl2018ffjord,song2020score} and use the RK45 ODE solver \citep{dormand1980family} in $\texttt{scipy.integrate.solve\_ivp}$ with $\texttt{atol=1e-5}$ and $\texttt{rtol=1e-5}$. To avoid numerical issues in practice, we set the limits of integration to be $(1,1\mathrm{e}{-5})$.


\begin{table}[H]
\centering
\begin{tabular}{c|c}
\hline
\textbf{Name}& \textbf{Component}\\
\hline
\textbf{Encoding Block} \\
\hline
conv1 & $3\times3$ conv, 64 filters, stride 1, bias=False \\
\hline
Dense Activation Block 1 & input dim=256, output dim=64 \\
\hline
Group Normalization 1 & num groups=4, num channels=64\\
\hline
conv2 & $3\times3$ conv, 128 filters, stride 2, bias=False \\
\hline
Dense Activation Block 2 & input dim=256, output dim=128 \\
\hline
Group Normalization 2 & num groups=32, num channels=128\\
\hline
conv3 & $3\times3$ conv, 256 filters, stride 2, bias=False \\
\hline
Dense Activation Block 3 & input dim=256, output dim=256 \\
\hline
Group Normalization 3 & num groups=32, num channels=256\\
\hline
conv4 & $3\times3$ conv, 512 filters, stride 2, bias=False \\
\hline
Dense Activation Block 4 & input dim=256, output dim=512 \\
\hline
Group Normalization 4 & num groups=32, num channels=512\\
\hline
\textbf{Decoding Block} \\
\hline
tconv4 & $3\times3$ 2d convtranspose, 128 filters, stride 2, bias=False \\
\hline
Dense Activation Block 5 & input dim=256, output dim=256 \\
\hline
Group Normalization 5 & num groups=32, num channels=256\\
\hline
tconv3 & $3\times3$ 2d convtranspose, 128 filters, stride 2, bias=False \\
\hline
Dense Activation Block 6 & input dim=256, output dim=128 \\
\hline
Group Normalization 6 & num groups=32, num channels=128\\
\hline
tconv2 & $3\times3$ 2d convtranspose, 64 filters, stride 2, bias=False \\
\hline
Dense Activation Block 7 & input dim=256, output dim=256 \\
\hline
Group Normalization 7 & num groups=32, num channels=64\\
\hline
tconv1 & $3\times3$ 2d convtranspose, 1 filter, stride 1 \\
\hline
Fully Connected Layer & input dim=784, output dim =1 \\
\hline
\end{tabular}
\caption{Convolutional U-Net architecture used for the energy-based modeling experiments for MNIST.}
\label{table:unet_arch}
\end{table}

\subsection{Additional Experimental Results}
\label{addtl-results}
\subsubsection{Synthetic Experiments}
\label{addtl-synthetic-app}
\paragraph{1-D Gaussians.}
As a warm-up, we evaluated whether the joint score matching objective in Section~\ref{joint_training} is able to recover the true log-ratios of two 1-dimensional Gaussian distributions. We experimented with $p(\rvx) = \mathcal{N}(0,1)$ and $q(\rvx) = \mathcal{N}(0,\sigma^2)$ on two tasks of increasing difficulty, where $\sigma^2 = 1$ and $\sigma^2 = 1\mathrm{e}{-6}$ respectively.
We use the arithmetic interpolation scheme of $\rvx(t) = \sqrt{1-t^2}\rvy + t \cdot \rvx$ for $\rvx \sim q(\rvx)$ and $\rvy \sim p(\rvx)$, and use SSM to learn the data scores.
Note that because both $p$ and $q$ are Gaussian, the form of the intermediate densities $p_t(\rvx) = \mathcal{N}(0, 1-(1-\sigma^2) t^2)$ can be obtained analytically.

Because the discrepancy between $p$ and $q$ is extremely large in these settings, we follow \citep{rhodes2020telescoping} and endow the score network with the true parametric forms of $\nabla_\rvx \log p_t(\rvx)$ and $\dt \log p_t(\rvx)$.
This way, the score network has to recover a single scalar parameter $\theta \in \bbR$. 
Specifically, we have: 
$$\nabla_\rvx \log p_t(\rvx) = \frac{-(\rvx-\theta)}{(1-[1-\sigma^2] \cdot t^2)}$$
$$\dt \log p_t(\rvx) = \frac{\left[-(\theta - x)^2 - (1-\sigma^2) \cdot t^2 + 1 \right]\cdot t(1-\sigma^2)}{(1-[1-\sigma^2] \cdot t^2)^2}$$

\begin{figure*}[!ht]
    \centering
        \subfigure[$p(\rvx) = \mathcal{N}(0,1)$ and $q(\rvx) = \mathcal{N}(0,0.01)$]{\includegraphics[width=.45\textwidth]{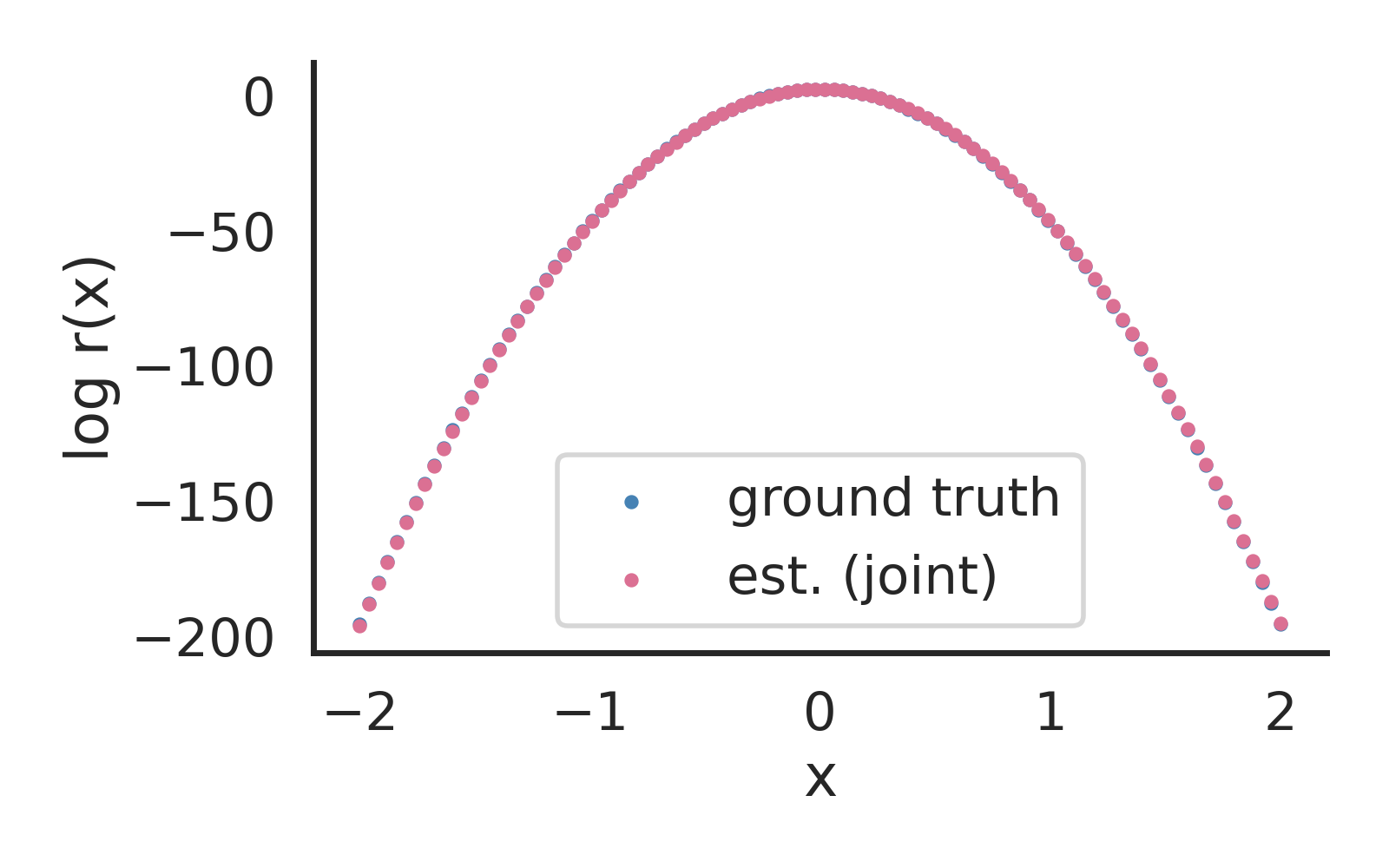}}
        \subfigure[$p(\rvx) = \mathcal{N}(0,1)$ and $q(\rvx) = \mathcal{N}(0,1\mathrm{e}{-6})$]{\includegraphics[width=.46\textwidth]{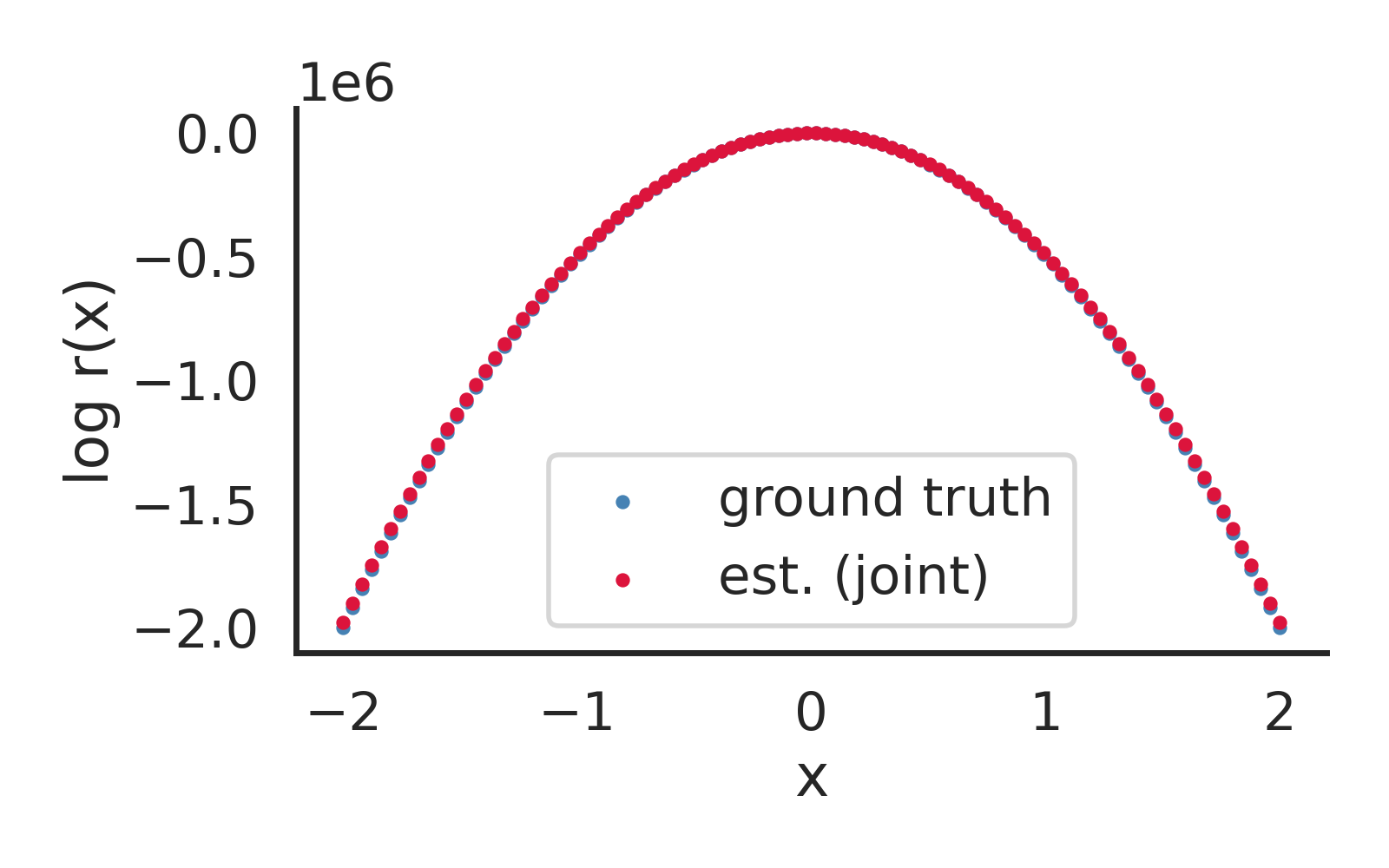}}
    \caption{Synthetic 1-D Gaussian example, demonstrating that the parameterized joint score network trained with SSM is able to recover the ground truth density ratios.}
    \label{fig:1d_exp}
\end{figure*}

Using this parameterization, we train the score model with the Adam optimizer with a learning rate of $0.001$ for 10,000 steps using a batch size of 128. As shown in Figure~\ref{fig:1d_exp}, we find that the joint score network is able to recover the true $\theta^*$. 

\paragraph{2-D Gaussians.}
In this setup, we remove the parameterization of the score network and directly learn the scores from data. We use fully-connected MLPs for all methods, including: (a) \texttt{NCE} (a single binary classifier); (b) TRE with 4 bridge distributions (\texttt{TRE(4)}); (c) TRE with 9 bridge distributions (\texttt{TRE(9)}); (d) the Time-only score network; and (e) the Joint score network. We use the VPSDE interpolation schedule for all methods.

For both TRE models, we found that naively using a deeper network hurt performance.
Therefore, we used a single input layer and single hidden layer both of size \texttt{z\_dim}=256 that were first used to transform all input features (and thus were shared across all bridges). This kind of parameter sharing was reported to be helpful in \citep{rhodes2020telescoping}. We then used 1 linear classification head per intermediate density. We found that \texttt{LeakyRELU} activations with coefficient 0.3 worked the best.

For the time score model, we used an MLP with \texttt{ELU} activations with 2 hidden layers. (input dimension + 1 because we concatenate the minibatch of sampled times to the data).
For the joint score network, we used an MLP with a single input layer and a single hidden layer that output twice the number of usual output features, then then split the output features into 2 blocks (one for the time scores, and another for the data scores). The time score-specific head was an MLP with 2 hidden layers, and the data-score specific head was an MLP with 2 hidden layers. 
The baseline classifier was an MLP with 2 hidden layers, the same as the time score network. 

In terms of hyperparameters, we swept through batch size=\{128,256\}, learning rate=\{2e-4,5e-4,1e-3\}, \texttt{z\_dim}=\{128,256\} and used the best model configurations for all methods.

We summarize the method-specific model architectures below:
\begin{enumerate}
    \item \textbf{NCE:} \texttt{Linear(2, 256) $\rightarrow$ \texttt{ELU} $\rightarrow$ Linear(256, 256) $\rightarrow$ \texttt{ELU} $\rightarrow$ Linear(256, 256) $\rightarrow$ \texttt{ELU} $\rightarrow$ Linear(256, 1)}.
    \item \textbf{TRE:} \texttt{Linear(3, 256) $\rightarrow$ \texttt{LeakyReLU(0.3)} $\rightarrow$ Linear(256, 256) $\rightarrow$ \texttt{LeakyReLU(0.3)} $\rightarrow$ [Linear(256, 1) for \_ in range(num\_bridges)]}.
    \item \textbf{Time:} \texttt{Linear(3, 256) $\rightarrow$ \texttt{ELU} $\rightarrow$ Linear(256, 256) $\rightarrow$ \texttt{ELU} $\rightarrow$ Linear(256, 256) $\rightarrow$ \texttt{ELU} $\rightarrow$ Linear(256, 1)}.
    \item \textbf{Joint (Shared):} \texttt{Linear(3, 256) $\rightarrow$ \texttt{ELU} $\rightarrow$ Linear(256, 512) $\rightarrow$ chunk(2) $\rightarrow$}
    \begin{enumerate}
        \item \textbf{Time Module:} \texttt{Linear(256, 256) $\rightarrow$ \texttt{ELU} $\rightarrow$ Linear(256, 256) $\rightarrow$ \texttt{ELU} $\rightarrow$ Linear(256, 1)}
        \item \textbf{Data Module:}  \texttt{Linear(256, 256) $\rightarrow$ \texttt{ELU} $\rightarrow$ Linear(256, 256) $\rightarrow$ \texttt{ELU} $\rightarrow$ Linear(256, 2)}
    \end{enumerate}
\end{enumerate}

\begin{figure*}[!t]
    \centering
        \subfigure[TRE(4) MSE: 1.1]{\includegraphics[width=.25\textwidth]{figures/2d_gmm_vpsde/TRE4_pq.png}}
        \subfigure[TRE(9) MSE: 0.63]{\includegraphics[width=.24\textwidth]{figures/2d_gmm_vpsde/TRE9_pq.png}}
        \subfigure[Time MSE (Ours): 0.58]{\includegraphics[width=.24\textwidth]{figures/2d_gmm/TimeScores.png}}
        \subfigure[Joint MSE (Ours): 0.35]{\includegraphics[width=.25\textwidth]{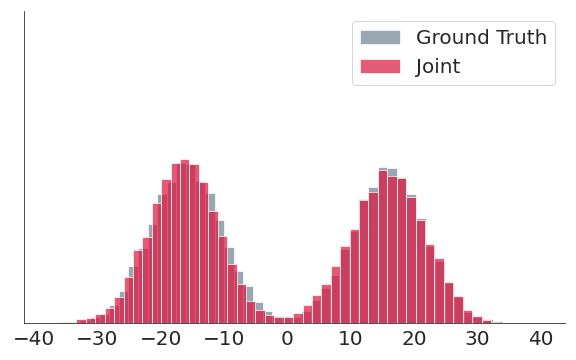}}
    \caption{Motivating example on a synthetic 2-D Gaussian dataset, with learned density ratio estimates by method relative to the ground truth values for (a-d). The performance of TRE improves with more intermediate bridge distributions, while our score matching method outperforms the rest. 
    The x-axis denotes the log-ratios.}
    \label{fig:2d_exp_supp}
\end{figure*}

Additionally, we incorporate results for the pathwise method on the original $p(\rvx) = \mathcal{N}(\bm{0}, \bm{I})$ and $q(\rvx) = \mathcal{N}(\bm{4}, \bm{I})$ setup as in Figure~\ref{fig:2d_exp}, rather than the more challenging evaluation setup in the main text with $q(\rvx) = \mathcal{N}(\bm{5}, \bm{I})$. We omit results for the naive baseline for clarity (and its poor performance). As shown in Figure~\ref{fig:2d_exp_supp}(d), we find that the pathwise method outperforms all other methods as expected.

\subsubsection{Mutual Information (MI) Estimation}
\label{appendix_mi}
For MI estimation between two high-dimensional correlated Gaussians, we follow the setup of \citep{rhodes2020telescoping} and parameterize the score network such that it only needs to learn a single $d \times d$ matrix (corresponding to $\bm{S}$ below). 
Similar to the 1-D Gaussians experiment, we use the arithmetic interpolation scheme to sample data points from all the intermediate densities. 
Concretely, we note that $p(\rvx) = \mcal{N}(\bm{0}, \bm{I})$ and $q(\rvx) = \mcal{N}(\bm{0}, \bm{\Sigma})$. Due to the way that we interpolate, the intermediate distributions are $p_t(\rvx) = \mathcal{N}(\bm{0}, \bm{I} + t^2(\bm{\Sigma} - \bm{I})) = \mathcal{N}(\bm{0}, \bm{I} + \bm{S} t^2)$, where we let $\bm{S} = (\bm{\Sigma} - \bm{I}) \in \bbR^{d \times d}$ for notational convenience. Then, our joint score network is trained to output:
$$\nabla_\rvx \log p_t(\rvx) = - \bm{M} \rvx$$
$$\dt \log p_t(\rvx) = -t \cdot \textrm{tr}(\bm{S} \cdot \bm{M}) + \rvx^\top \bm{M} \bm{S} \bm{M} \rvx$$
where $\bm{M} = (1 + t^2 \bm{S})^{-1}$.

For $d=\{40,80,160\}$, we use batch sizes of 512 and use a batch size of 256 for $d=320$. We use the Adam optimizer with learning rate 0.001 with weight decay of 0.0005 for all settings. We train for $\{30K, 50K, 200K, 200K\}$ steps for $d=\{40,80,160,320\}$ respectively.
We note that we used a heuristic weighting function that led to good performance for this experiment. Specifically, we let $\lambda(t) = (1-t^2)$ due to the way that we construct the intermediate samples $\rvx(t)$. 

For TRE, we used their default hyperparameter settings and architecture details and refer the reader to \citep{rhodes2020telescoping} for more details on the exact experimental setup. In terms of the number of intermediate densities, we used $\{2,4,6,8\}$ bridge distributions for $d=\{40,80,160,320\}$ respectively.

\subsubsection{Energy-Based Modeling with MNIST}
\label{app-mnist}
Our experimental setup largely mirrors that of \citep{rhodes2020telescoping} and refer the reader to their paper for additional details.

\paragraph{Data Preprocessing.} To match the experimental setting as closely as possible, we first rescale the pixels to lie in $[0,1]$, apply uniform dequantizatiation, and logit-transform the dequantized pixel values. Then, we whiten the transformed dataset by subtracting off the mean before training our flow models. For training the score network on the data distribution $q(\rvx)$, we rescale the pixel values to lie between $[-1, 1]$.

\paragraph{Fitting noise distributions.} We experiment with three different interpolation schemes, which first require training a ``normalizing flow'' (invertible transformation) for each setting on the MNIST \citep{lecun1998mnist} dataset. That is, our prior distribution $p_1(\rvx)$ is the density captured by a pretrained flow on MNIST, and we can utilize the flow mapping to interpolate in the latent space $p(\rvz) = \mathcal{N}(\bm{0}, \bm{I})$. We found this latent interpolation procedure to be critical for the success of all methods involved in this experiment.

\begin{figure*}[!t]
    \centering
        \subfigure[Gaussian: 2.01 bpd]{\includegraphics[width=.32\textwidth]{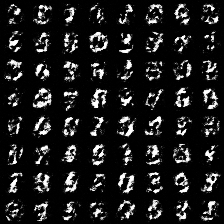}}
        \subfigure[Gaussian Copula: 1.44 bpd]{\includegraphics[width=.32\textwidth]{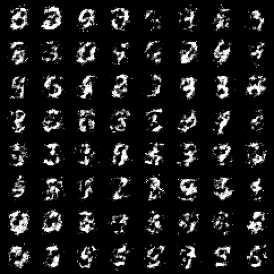}}
        \subfigure[RQ-NSF: 1.12 bpd]{\includegraphics[width=.32\textwidth]{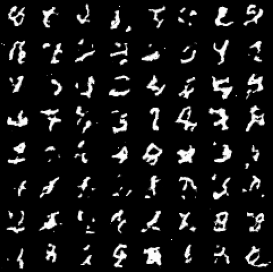}}
    \caption{Samples from the transformed noise distributions $p(\rvx)$ in the energy-based modeling experiments for MNIST: (a) Gaussian model (an affine transformation); (b) Gaussian copula parameterized by the Rational Quadratic Spline building block; (c) the RQ-NSF flow.}
    \label{fig:flow_samples}
\end{figure*}

For the copula, we use a batch size of 512 and train for 40K iterations. For both the copula and the RQ-NSF, we use a multi-scale convolutional neural network (CNN) with 2 levels, where each level contains 8 steps. The coupling transofrms use 64 feature maps and the spline functions use 8 bins with the interval width between [-3, 3]. We use a learning rate of 0.0001 and use a cosine annealing decay schedule. For the RQ-NSF, we use a learning rate of 0.0005 and train for 200,000 steps with a batch size of 256. This allowed us to match the initial Noise Distribution bpds of \citep{rhodes2020telescoping} with the exception of the Gaussian copula, where we were unable to improve upon a bpd of 1.44. However, we note that our method was still able to outperform the TRE baseline in Section~\ref{sec:mnist}. Samples from all noise distributions are shown in Figure~\ref{fig:flow_samples}.

\paragraph{Training the score networks.} For training the score networks, we perform a hyperparameter sweep where \texttt{lr}=\{2e-4, 5e-4, 1e-3\}, \texttt{batch\_size}=\{128, 256, 500\}, and we use the loss history as described in Section~\ref{app:loss_history}. For the interpolation mechanism, we use the VPSDE noise schedule in latent space. For evaluating likelihoods, we use EMA with rate=0.999. For the score network architecture, we use a convolutional U-Net \citep{ronneberger2015u} with a smaller MLP for the time embeddings.
For the RQ-NSF setting, the powerful flow network made it challenging for the score network to make additional improvements on the likelihoods. For this model, we did not use the loss history, and directly reweighted the loss using the VPSDE reweighting scheme, which performed the best out of all configurations. 

\begin{figure*}[!t]
    \centering
        \subfigure[Gaussian: 1.33 bpd]{\includegraphics[width=.32\textwidth]{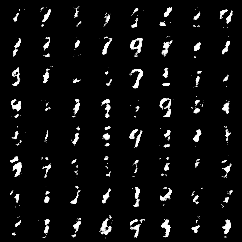}}
        \subfigure[Gaussian Copula: 1.21 bpd]{\includegraphics[width=.32\textwidth]{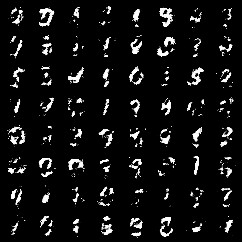}}
        \subfigure[RQ-NSF: 1.08 bpd]{\includegraphics[width=.32\textwidth]{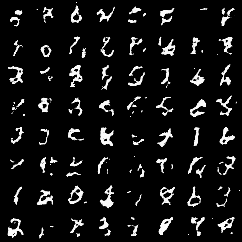}}
    \caption{Samples obtained from running AIS with 100 parallel chains for 1000 steps for the energy-based modeling experiments for MNIST: (a) Gaussian model (an affine transformation); (b) Gaussian copula parameterized by the Rational Quadratic Spline building block; (c) the RQ-NSF flow.}
    \label{fig:ais_samples}
\end{figure*}


\paragraph{AIS and likelihood evaluation}
For likelihood evaluation, we computed the bpds directly from the score network and also via AIS. We ran AIS with 100 parallel chains for 1000 steps. We used Hamiltonian Monte Carlo (HMC) for MCMC, where we conducted the sampling in z-space and mapped the results back to x-space. Samples obtained from running AIS are shown in Figure~\ref{fig:ais_samples}.

\subsubsection{Exploration of DRE-$\infty$'s computational gains}
We note that TRE’s ResNet architecture did not perform well for score estimation (and vice versa), so we used smaller U-Nets \citep{salimans2021should}. On MNIST with the original TRE codebase, TRE with 10 bridges (“TRE-10”) has 7.4M trainable parameters, while our time score network has 3.7M parameters. TRE-30, on the other hand, has 19.2M parameters. However, we do need to train longer to converge -- while TRE converges in about a day, our models took roughly 2 days. We expect improvements in
optimization as well as variance reduction to accelerate training. 

We explore how many time score evaluations typically occur while performing the numerical integration at test time. 
As expected, the average number of function evaluations varies with the difficulty of the task. 
For example on the MNIST dataset, the score network trained with the Gaussian noise model requires $266.6 \pm 13.3$ function
61 evaluations at an error tolerance of $1e5$, $199.4 \pm 5.7$ evaluations for the copula noise model, and $118.4 \pm 2.9$ evaluations for the RQ-NSF.

In terms of wall-clock time, this helps our approach perform favorably against TRE when evaluating the log-ratios after training. For the RQ-NSF noise model for a batch of 100 examples, TRE-10 took $1.26 \pm .02$s, TRE-15 took $1.42 \pm .07$s, TRE-30 took $2.01 \pm 0.27$s, and ours took $0.70 \pm 0.02$s at error tolerance $1e5$.



\subsection{Societal Impact}
Ultimately, the goal of this work is to provide more accurate density ratio estimates for a wide variety of machine learning applications. While $\dreinf$ in itself does not have any direct social implications, its use in downstream applications such as domain adaptation, anomaly detection, and propensity score matching in causal inference, etc. may have consequences depending on their use case.



\end{document}